\documentclass[final,  3p, times]{elsarticle}

\usepackage{lineno}
\modulolinenumbers[5]

\usepackage{graphicx}
\usepackage{amsmath}
\usepackage{amssymb}
\usepackage{booktabs}
\usepackage{verbatim}
\usepackage{subcaption}
\usepackage{mathtools}

\usepackage{url}
\usepackage{times}
\usepackage{epsfig}
\usepackage{subcaption}
\usepackage{multirow}
\usepackage{comment}
\usepackage{balance}
\usepackage{physics}
\usepackage{arydshln}

\usepackage{algorithm,algpseudocode}
\usepackage{amsfonts}

\graphicspath{ {imgs/} }

\usepackage{dsfont}
\usepackage{mathtools}
\usepackage{booktabs}
\usepackage{graphics}
\usepackage[mathscr]{euscript}

\renewcommand{\vec}[1]{{\mathbf #1}}
\newcommand{\mat}[1]{{\mathbf #1}}

\DeclareMathOperator*{\E}{\mathbb{E}}

\usepackage[pagebackref,breaklinks,colorlinks]{hyperref}

\usepackage[normalem]{ulem}
\usepackage{xspace}
\usepackage{dsfont}
\usepackage[accsupp]{axessibility}

\usepackage{xcolor,colortbl}

\definecolor{Gray}{gray}{0.9} 

\usepackage{pifont}
\newcommand{\cmark}{\ding{51}}%
\newcommand{\xmark}{\ding{55}}%

\def\onedot{\ifx\@let@token.\else.\null\fi\xspace}
\def\eg{\emph{e.g}\onedot} 
\def\ie{\emph{i.e}\onedot}

\def\etal{\emph{et al}\onedot}

\makeatother

\usepackage[capitalize]{cleveref}
\crefname{section}{Sec.}{Secs.}
\Crefname{section}{Section}{Sections}
\Crefname{table}{Table}{Tables}
\crefname{table}{Tab.}{Tabs.}

\journal{Neural Networks}

\begin{document}

\begin{frontmatter}

\title{Subspace Distillation for Continual Learning}

\author[1,3]{Kaushik Roy\corref{c1}}
\ead{kaushik.roy@{monash.edu, csiro.au}}
\author[1,2,3]{Christian Simon\corref{c1}}
\ead{christian.simon@anu.edu.au}
\author[3,4]{Peyman Moghadam\corref{c1}}
\ead{peyman.moghadam@{csiro.au, qut.edu.au}}
\author[1,3]{Mehrtash Harandi\corref{c1}}
\ead{mehrtash.harandi@monash.edu}

\cortext[c1]{Corresponding author}

\address[1]{Monash University; Melbourne, VIC, Australia}
\address[2]{Australian national University; Canberra, ACT, Australia}
\address[3]{CSIRO, Data61; Brisbane, QLD, Australia}
\address[4]{Queensland University of Technology; Brisbane, QLD, Australia}

\renewcommand*{\thefootnote}{\fnsymbol{footnote}}
\footnotetext[0]{Manuscript submitted in May 2022. Manuscript accepted by Neural Networks in July 2023.}

\begin{abstract}

An ultimate objective in continual learning is to preserve knowledge learned in preceding tasks while learning new tasks. 
To mitigate forgetting prior knowledge, we propose a novel knowledge distillation technique that takes into the account the manifold structure of the latent/output space of a neural network in learning novel tasks. To achieve this, we propose to approximate the data manifold up-to its first order, hence benefiting from linear subspaces to model the structure and  maintain the knowledge of a neural network while learning novel concepts.
We demonstrate that the modeling with subspaces provides several intriguing properties, including robustness to noise and therefore effective for mitigating Catastrophic Forgetting in continual learning.
We also discuss and show how our proposed method can be adopted to address both classification and segmentation problems.
Empirically, we observe that our proposed method outperforms various continual learning methods on several challenging datasets including Pascal VOC, and Tiny-Imagenet. Furthermore, we show how the proposed method can be seamlessly combined with existing learning approaches to improve their performances.
The codes of this article will be available at \url{https://github.com/csiro-robotics/SDCL}.

\end{abstract}

\begin{keyword}
Lifelong Learning \sep Subspace Distillation \sep Knowledge Distillation \sep Continual Semantic Segmentation \sep  Catastrophic Forgetting \sep Background Shift \sep Continual Learning
\end{keyword}

\end{frontmatter}

\section{Introduction}
\label{sec:intro}

\underline{C}ontinual \underline{L}earning (CL) is the process of robust, efficient and gradual learning in non-stationary environments. A fundamental aspect of intelligence is the capability of incrementally learning from sequential experiences. 
Equipping neural networks with CL capability requires the model to preserve its previously learned experiences while acquiring novel knowledge. Neural networks, trained in an offline mode%
, are currently the method of choice in a wide spectrum of problems in AI and machine learning. The underlying assumption here is that the model has the knowledge about all the decisions it should take in the future apriori. For example, all classes a model will encounter in future are known in an image classification task.
Furthermore, in offline training, data used for training the model in future steps should be i.i.d, otherwise internal representations learned by the model are hardly useful. 

In this paper, our focus is to design a mechanism that enables neural network model to learn continually in a dynamic environment.
One may wonder \emph{is it advantageous for a model to learn sequentially like humans?} Continual learning techniques will endow our machines to learn potentially  over a lifetime, as does a human. Furthermore, having visual understanding and semantic segmentation in mind, continual adaptation to a changing target specification enables the model to learn a diverse, and growing set of  classes. This aspect of continual learning is commonly considered as a necessity towards human-level artificial general intelligence~\cite{hadsell2020embracing}. Also, we note that  continual learning methods could offer profound advantages for models even in stationary settings, by enabling them to improve their efficacy without the need to train from scratch upon availability of new data.

In a continual learning setting, current \underline{D}eep \underline{N}eural \underline{N}etworks (DNNs) exhibits a drastic fall in the overall performance when the model is trained on a series of tasks. Precisely, in absence of samples from old tasks, the performance degrades on previously encountered tasks after the model is trained on a novel tasks. In other words, the knowledge from previous tasks gets overwritten thoroughly and DNNs forget  previously learned tasks abruptly once information relevant to novel tasks is presented~\cite{parisi2019continual}. This phenomenon of forgetting prior tasks because of the changes on critical weights related to previously observed tasks is often referred to as \textbf{Catastrophic Forgetting}~\cite{goodfellow2013empirical, kirkpatrick2017overcoming, robins1995catastrophic, french1999catastrophic, thrun1998lifelong}. Therefore, to design DNNs for \underline{C}ontinual \underline{L}earning, one needs to address Catastrophic Forgetting.

\begin{figure}[]
    \centering
        \includegraphics[width=.75\columnwidth]{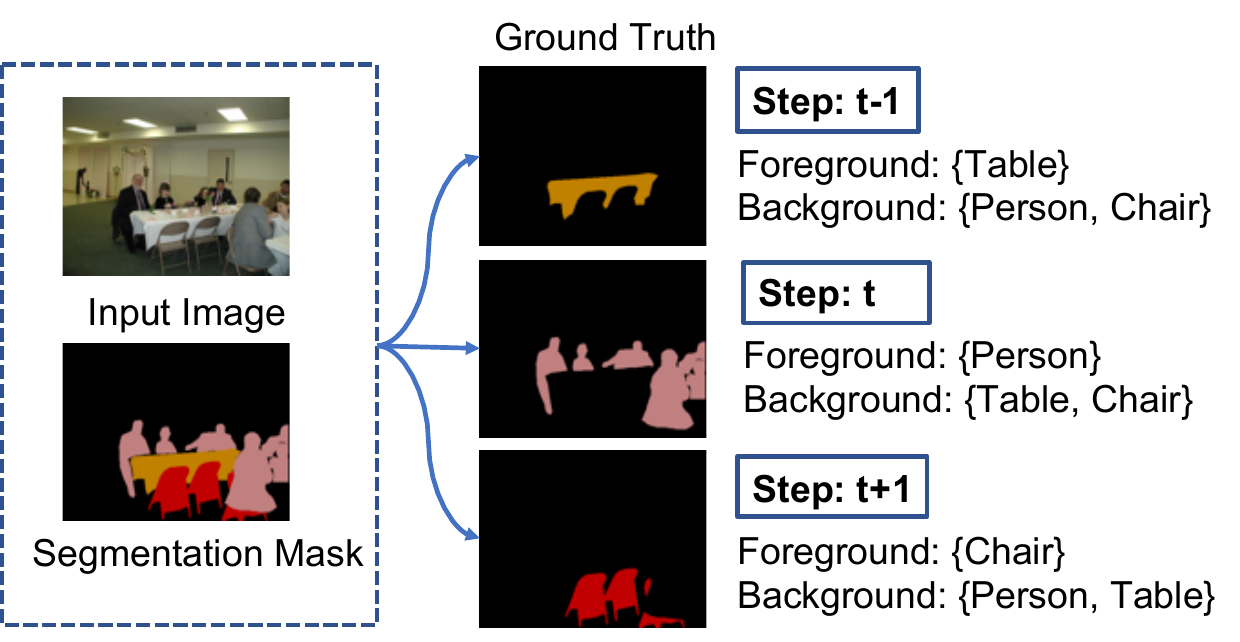}
        \caption{Semantic shift of background classes at different learning steps in Continual Semantic Segmentation (CSS). Classes (\eg. table and chair) from the old task at step $t$  and future task at step $t+1$ are collapsed into background at current task $t$.}
        \label{fig:bg_shift}
\end{figure}

Another problem of interest in this paper is \underline{C}ontinual \underline{S}emantic \underline{S}egmentation (CSS). Semantic segmentation~\cite{chen2018encoder, long2015fully, badrinarayanan2017segnet} is the task of  assigning a category label such as ``person'' or ``vehicle'' to  every single pixel of an image. In class-incremental CSS problem, a model is sequentially exposed to learn a set of novel classes. At the end of each training step, the CSS model is supposed to classify a pixel with all the seen classes until current task for evaluation. Aside from catastrophic forgetting, in CSS, we need to tackle another fundamental problem, namely \textbf{Background Shift}~\cite{cermelli2020modeling}.

In a conventional semantic segmentation setup, all object categories are predefined, and the class ``background'' encapsulates all other object categories that are not relevant to the problem at hand.
In contrast and in CSS, at each learning step, the class background merely corresponds to categories that do not belong to any of the classes at the current step (see Fig.~\ref{fig:bg_shift}).
As a result, the class \emph{background} contains not only pixels from \emph{unseen} and \emph{future classes} but also pixels from previously seen and old classes. 
This setting can be considered as a dense prediction task with noisy labels as the future unseen classes or old seen ones are grouped under a super-class named \emph{background}. If certain measures are not taken, the background shift could exacerbate the catastrophic forgetting even further.

\begin{figure*}[htbp]
\vspace{-2mm}
    \centering
        \includegraphics[width=\textwidth]{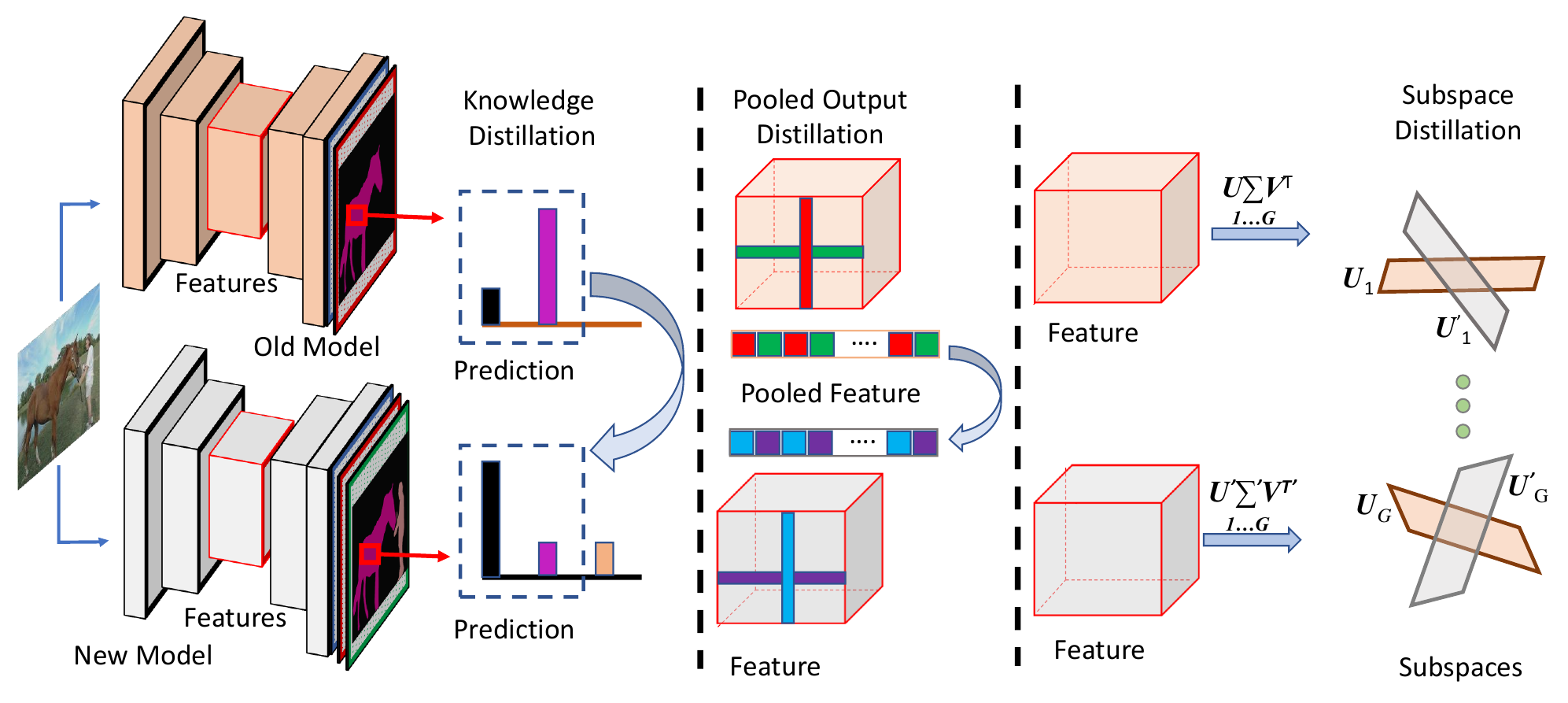}
        \caption{Distillation strategies: (Left) Knowledge Distillation works on output layer and matches the probability distribution between old and new model, (Middle) Pooled Output Distillation (POD) applies similarity constraints on the pooled feature between old and new model and (Right) Subspace Distillation on the other hand maps corresponding subspaces constructed from the intermediate feature maps from old and new model. }
        \label{fig:distillation}
\end{figure*}

A common way of addressing Catastrophic Forgetting and Background Shift is to distill the knowledge from old model to current one (see Fig.~\ref{fig:distillation}). Distillation methods such as LwF~\cite{li2017learning}, PODnet~\cite{douillard2020podnet}), often match the output/latent representation of a network, and hence ensuring that the prior knowledge remains unchanged in current model and the performance remains consistent on old tasks. %

In this paper, we introduce a structured form of knowledge distillation~\cite{hinton2015distilling} that is suitable for both \underline{C}ontinual \underline{L}earning with class-incremental setting and  \underline{C}ontinual \underline{S}emantic \underline{S}egmentation (CSS)~\cite{michieli2019incremental, cermelli2020modeling}. Precisely, we propose to distill the structure of the feature space in the intermediate layers of neural network to preserve the previously learned knowledge in the current model. 
\begin{figure}[]
    \centering
        \includegraphics[width=.45\columnwidth]{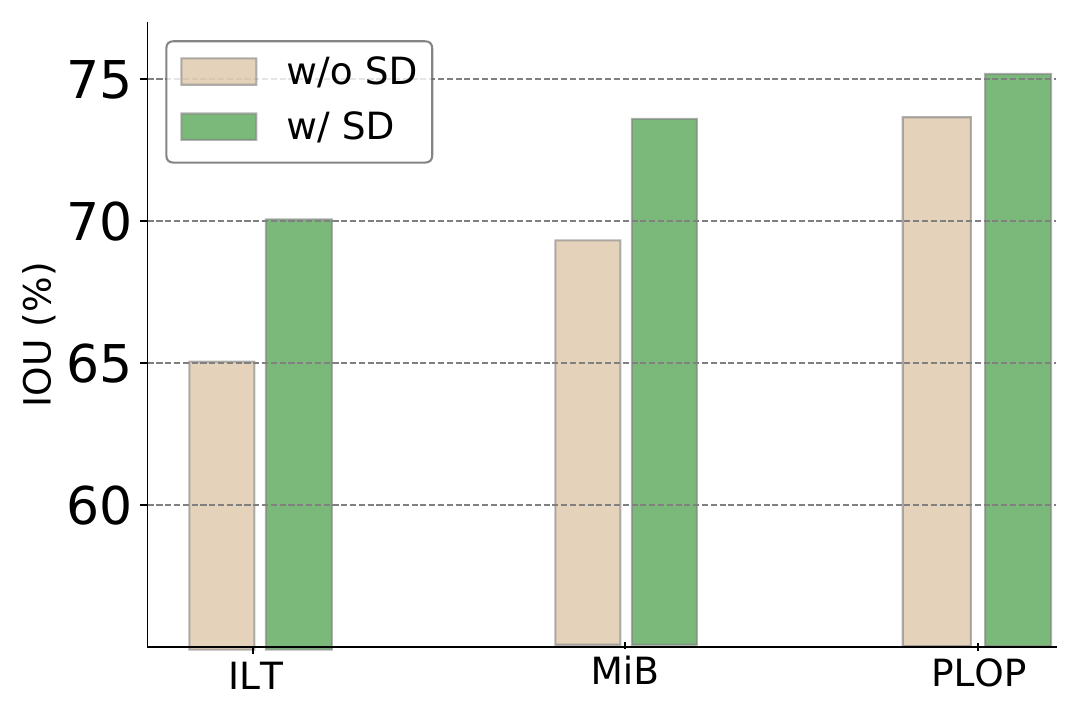}
        \caption{%
        Average IOU at the end of training for 19-1 task setting on Pascal VOC dataset using different CSS methods with or without our proposed Subspace Distillation (SD). We see a significant improvement of IOU when Subspace Distillation is added with output distillation based methods: ILT~\cite{michieli2019incremental}, and MiB~\cite{cermelli2020modeling} and pseudo-labeling based feature distillation method:  PLOP~\cite{douillard2021plop} methods.%
        }
        \label{fig:iou_improvement}
\end{figure}
Our proposed method encodes the structure via low-dimensional subspaces. Subspaces have been used in a broad range of problems in computer vision to model the data manifold locally. Subspaces are robust to perturbation, and can be computed for high-dimensional data easily, hence has been employed with success for adapting neural networks~\cite{simon2020adaptive, zhang2019neural}. %
Therefore, to mitigate catastrophic forgetting in CL, we propose to maintain geometric structure of the feature space through encoding subspaces between models from sequential learning steps.
Our approach starts with decomposing extracted feature map in the intermediate layers of deep neural network followed by constructing structures of it through selecting prominent subspaces that approximate the data manifold to the first-order. This enables us to impose constraint to maintain similar subspace structures between old and new models. 
In our method, we formulate the constraints using the geometry of Grassmannian~\cite{edelman1998geometry}, and propose to minimize the distance between corresponding feature subspaces of old and current model.

As a motivating example, to examine the ability of subspace distillation in improving catastrophic forgetting by preserving prior knowledge, %
we evaluate the performance of subspace distillation by adding it to existing state-of-the-art CSS methods, \eg. ILT~\cite{michieli2019incremental}, MiB\cite{cermelli2020modeling}, and PLOP~\cite{douillard2021plop} and report the result in Fig.~\ref{fig:iou_improvement}. The result shows that the average IOU for all of the three methods improves significantly on the Pascal VOC dataset. %
Furthermore, our proposed subspace distillation algorithm can be merged with other distillation techniques seamlessly and provides them with complementary constraints to enforce structural similarities.

Overall, our contributions in this paper are as follows:
\begin{itemize}
  \item We propose a robust feature distillation strategy, namely Subspace Distillation (SD) to tackle catastrophic forgetting in CL through applying constraint on maintaining similar feature structure between old and new model.
  \item We present a generalized end-to-end continual learning framework using our proposed Subspace Distillation (SD) strategy in presence of a small subset of already observed samples from past tasks. %
  \item Our proposed Subspace Distillation (SD) strategy requires backpropagation through Singular Value Decomposition (SVD) method as it relies on SVD to compute the basis of subspaces. We show how this can be done in closed form solution by using partial derivative.
  \item Our proposed approach outperforms state-of-the-art continual learning methods on MNIST, CIFAR10 and Tiny-Imagenet datasets with varying memory size.
  \item We also show that a significant improvement can be achieved by combining subspace distillation strategy with existing methods for CSS on Pascal VOC dataset for different short and long task settings.
\end{itemize}

\section{Related Work}
\label{sec:related_work}

In this section we discuss related works in class-incremental learning and continual semantic segmentation as the targeted application.

\subsection{Continual Learning}

A variety of methods has been proposed to alleviate catastrophic forgetting for class-incremental classification problems~\cite{kirkpatrick2017overcoming, li2017learning, douillard2020podnet, rolnick2019experience, mallya2018packnet, yoon2017lifelong}. %
These methods for continual learning are classified primarily into three categories, such as regularization, dynamic architecture and memory replay based methods ~\cite{ebrahimi2019uncertainty}.

\noindent \textbf{Regularization based methods} preserve already learned information by imposing constraints on the update of weight~\cite{aljundi2018memory, chaudhry2018riemannian, kirkpatrick2017overcoming, zenke2017continual}, intermediate feature representation~\cite{douillard2020podnet, hou2019learning, dhar2019learning}, prediction~\cite{li2017learning, rebuffi2017icarl}, gradient~\cite{lopez2017gradient, chaudhry2018efficient}, or combination thereof.  Li \etal{}~\cite{li2017learning} in Learning without Forgetting (LwF) introduced knowledge distillation strategy in the output layer to minimize the dissimilarity between old task and new one. The significance of each synapses is measured to penalize update on most influential synapses and a surrogate loss function to estimate loss for old tasks is used with modified cost function in SI~\cite{zenke2017continual}. %
Hou \etal in~\cite{hou2019learning} proposed cosine normalization to combat catastrophic forgetting and facilitate the seamless integration of new and prior knowledge within a continual learning model. By adjusting the magnitudes of the model's weights, cosine normalization provides precise control over the influence of novel tasks. This mechanism ensures that the model does not prioritize the new task at the expense of previously learned tasks, thereby preserving valuable knowledge while accommodating novel knowledge. This normalization approach helps to mitigate the negative impact of catastrophic forgetting and enhances the model's ability to generalize across multiple tasks. Cheraghian~\etal in~\cite{cheraghian2021semantic} introduced a semantic-aware distillation loss for few-shot class incremental learning that takes into account the semantic structure of the data. The distillation strategy utilizes semantic embeddings associated with each class to guide the distillation process. The integration of semantic information fosters the preservation of prior knowledge in the continual learning (CL) model by facilitating learning not only from the representations of the previous model but also from the semantic relationships among the classes. During the incremental learning process, PODNet in~\cite{douillard2020podnet} used a Pooled Outputs Distillation (POD) mechanism to transfer knowledge from the previously learned tasks to the current task. Specifically, the outputs of the intermediate layers of the previous model are pooled and distilled into the corresponding layer of current model through minimizing the discrepancies using Euclidean distance of L2-normalized features. Mathematically, POD can be expressed as 
$L_{POD} = \left\|f_i^{old}(\vec{x}) - f_i^{new}(\vec{x})\right\|^2$,
where $f_i^{old}(\vec{x})$ and $f_i^{new}(\vec{x})$ are the pooled output of the $i^{th}$ spatial position for input $\vec{x}$ using old and new model respectively.
One of the limitations of PODNet is the absence of an explicit mechanism to effectively preserve the underlying latent structure, as it primarily relies on imposing constraints on the pooled features. Consequently, the presence of outliers or noisy attributes can potentially compromise the effectiveness of distillation in PODNet. In contrast, our proposed subspace distillation approach addresses this concern by imposing constraints on the low-dimensional subspaces derived from the latent features of both the old and current models. This not only enhances robustness to noise but also ensures the preservation of latent structure in continual learning scenarios.

\noindent \textbf{Dynamic architecture based methods} allocate new neurons to adapt to novel task. Andrei \etal{}~\cite{rusu2016progressive} introduced progressive network where old knowledge remains unchanged by keeping previously trained model frozen and a novel sub-network with fixed resources is allocated to learn new knowledge. Yoon \etal{}~\cite{yoon2017lifelong} proposed dynamically extendable network (DEN) that learns a compact representation by selective training and expanding neural network capacity by optimal number of units when new task arrives. Recently, Douillard~\etal in~\cite{douillard2021dytox} proposed first transformer based architecture where dynamic expansion of task specific token is used.

\noindent \textbf{Memory-based methods} partially stores the previous data and train model by replaying stored old data together with new data~\cite{rebuffi2017icarl, gepperth2016bio}. Rebuffi~\etal{}~\cite{rebuffi2017icarl} introduced iCARL method where herding based sample selection was used to keep a small portion of previous dataset in the memory and replayed interleaved with new samples.
Many recent approaches have extended iCARL to bias correction problem for classifier~\cite{javed2018revisiting}, a metric learning model for imbalance dataset~\cite{hou2019learning}, a memory sampling method~\cite{aljundi2019gradient, aljundi2019online, isele2018selective}. %
Javed \etal{}~\cite{javed2018revisiting} introduced a dynamic threshold moving method to address the classifier bias generated by the knowledge distillation approach in iCARL. Hou~\etal~\cite{hou2019learning} in LUCIR proposed to combine inter-class separation constraint, old classes geometric structure preserving constraint and cosine normalization to tackle imbalance dataset problem. Aljundi~\etal~\cite{aljundi2019gradient} claimed that memory sampling method is crucial, and accuse random memory sample selection for sub-optimal performance on old tasks. In~\cite{aljundi2019gradient} replay memory sampling is defined as a constrained optimization problem and formulated as solid angle minimization problem to maximize the diversity in the replay memory. While in iCARL, the memory constitutes samples randomly chosen based on the cluster centers, in~\cite{aljundi2019online} and~\cite{isele2018selective}, diverse sample selection mechanisms, including most interfered sample retrieval and global distribution matching based sampling are utilized. %

To address the security concern, few recent approaches employed generative model to produce samples belonging to old classes~\cite{shin2017continual} instead of storing subset of old samples in memory. Deep Generative Replay (DGR)~\cite{shin2017continual} proposed a dual-model architecture, one for generating pseudo samples and another for solving tasks by replaying pseudo samples together with new samples. To reduce the memory footprint of storing real samples, compressed feature have been stored in~\cite{hayes2020remind, iscen2020memory}. REMIND~\cite{hayes2020remind} used product quantization method to quantize latent representation and stored indices in memory that were used later to decode the representation for reply. However, because of incremental update in model, stored latent representation also requires adaptation. To fit the stored representation into current latent space, Iscen~\etal ~\cite{iscen2020memory} employed a multi-layer perceptron to map corresponding old and new feature map generated for images of current task.%

Recently, Dark Experience Replay (DER++)~\cite{douillard2021tackling} proposed to store both logit, and label for corresponding sample and used the reservoir sampling strategy to select samples from data stream for memory buffer. In DER++, knowledge distillation is performed by mapping output logit from current model with corresponding memory logit.

\subsection{Continual Semantic Segmentation}

In recent time, a growing number of works have emerged into continual semantic segmentation. In the literature, CSS methods primarily fall into two major categories: (i) regularization and (ii) Replay based model.

Following the success of regularization methods in CL, several works have proposed mechanism for controlling update of neuron weights to mitigate catastrophic forgetting in CSS~\cite{michieli2019incremental, cermelli2020modeling, douillard2021plop, michieli2021continual}. ILT~\cite{michieli2019incremental} investigates a \textbf{regularization-based technique} by freezing weights of the encoder network after learning the first task and applying knowledge distillations for the upcoming tasks. %
ILT is also equipped with the masked cross-entropy loss and masks on the output of the current model to consider only the seen classes. However, this approach doesn't resolve the background shift problem of CSS properly. %
Since the background in CSS may include previously seen objects, MiB~\cite{cermelli2020modeling} proposed to preserve the knowledge of an old model with knowledge distillation. However, the difference compared to standard knowledge distillation is that the probabilities of the background and the novel classes are appended such that the non-overlapping prediction of a novel class as the output of a current model can be aligned with the output of a previous model.
As knowledge distillation shows a promising direction for CSS, PLOP~\cite{douillard2021plop} introduced pseudo-labelling to tackle the background shift problem and adapt Pooled Out Distillation (POD)~\cite{douillard2020podnet} to preserve the previously learned knowledge. Since POD is developed for a classification problem in classical continual learning settings with reliance on global statistics, thus %
PLOP uses a multi-scale version of POD to integrate local and global statistics at different intermediate layers. %
SDR~\cite{michieli2021continual} leveraged contrastive learning together with novel sparsity constraint and prototype matching strategy to efficiently learn novel tasks and mitigate forgetting of prior knowledge. In SDR, to organize geometric structure of feature representation, clusters of data are described using prototypes that are forced to be closer in consecutive learning steps and far apart from one another by using prototype matching and repulsive force. Additional sparsity constraint imposed on feature representation of same classes helped to construct well-separated and tight cluster as well as create space for accommodating novel classes. %

\noindent
\textbf{Replay based methods} tackle the catastrophic forgetting in CSS by either storing real images from past tasks~\cite{cha2021ssul} or generating synthetic data of prior classes~\cite{maracani2021recall}. Cha~\etal in SSUL~\cite{cha2021ssul} proposed to extract future classes from background that are defined as unknown classes to facilitate learning novel classes. SSUL used a subset of old samples  for the first time in the literature of CSS to improve stability and plasticity of model. Additionally, freezing weights of encoder and old classifiers with binary cross entropy loss, and pseudo-labeling techniques helped to improve catastrophic forgetting in~\cite{cha2021ssul}. Maracani~\etal~\cite{maracani2021recall} employed generative model to produce samples from previously seen tasks together with newly proposed background impainting method for pseudo-labeling to tackle background shift and catastrophic forgetting.

\section{Preliminaries}
\label{sec:preliminaries}

In continual learning, a model needs to learn from a series of tasks. Each learning task is represented by its training set, often samples from novel classes or concepts. 
Let $\mathds{T} = \{\mathcal{T}_{1}, \mathcal{T}_{2},\cdots,\mathcal{T}_{T} \}$ be a sequence of $T$ tasks. In the setup we are interested in this work, every task comprises of 
\[\mathcal{D}^t = \Big\{\big(\mat{X}_i,y_i\big)\Big\}_{i=1}^{n_t}\]
where $\mat{X}_i \in \mathcal{X}$ denotes a training image of size $W \times H$ and $y_i \in \mathcal{Y}$ is the corresponding class belonging to current task $t$. We also maintain a fixed size memory $\mathcal{M}$ to retain a small subset of samples from old tasks to better tackle the catastrophic forgetting. 
The goal of our knowledge distillation based approach is to apply constraints on updating weights of model so that the model generates similar latent representation and prediction in future tasks (\ie, $t, t+1, \cdots$) for what it has learned previously (\ie, $1, 2, \cdots, t-1$).

There is a subtle difference in setup when CL is considered for  semantic segmentation. For the problem of continual semantic segmentation, the training set consists of input samples and their corresponding segmentation mask. 
We denote the training set of CSS by
\[\mathcal{D}^t = \Big\{\big(\mat{X}_i,\vec{Y}_i\big)\Big\}_{i=1}^{n_t}\]
where $\vec{Y}_i \in \mathcal{Y}$ is the corresponding segmentation mask for the input image $\mat{X}_i$. Each pixel of the image belongs to a set of classes given by the current task $C^t$. Previously observed classes (\ie, $\mathbb{C}^{t-1}$) and future classes (\eg,  $\mathbb{C}^{t+1}$) are all labeled as the background class $c_{bg}$ for task $t$. In both continual semantic segmentation and classification tasks, at step $t$, the model should be able to predict all the observed classes, $\mathbb{C}^{1:t}$, throughout the learning experience.

\subsection{Evaluating an Incremental Learning Model}
\paragraph{\textbf{Class-Incremental Learning} The performance of continual learning methods are measured using the average task accuracy~\cite{chaudhry2018riemannian} in experiments.
Average accuracy is computed by average performance across all the previously observed and current tasks after training on current task $t$ and is defined as:
\begin{equation}
\mathrm{Acc}_{t} = \frac{1}{t}\sum _{i=1}^{t} \mathrm{Acc}_{t, i},
\label{eq:accuracy}
\end{equation}
where $Acc_{t, i}$ is the accuracy of task $i$ after learning task $t$.
}
\paragraph{\textbf{Continual Semantic Segmentation} Intersection Over Union (IOU)\cite{everingham2010pascal} metric is commonly used to evaluate the robustness towards catastrophic forgetting, in other words stability, and the ability of learning new class (\ie, plasticity) of CSS methods. IOU is computed at the end of learning all task at step $t$ for \textbf{(i)} initial set of classes $C^1$ at first task, (\textbf{ii)} incrementally leaned classes $C^{2:t}$, and \textbf{(iii) }all classes $C^{1 : t}$. IOU is defined as follows
\begin{align}
    \label{eqn:iou}
    \mathrm{IOU} =
    \dfrac{\mathrm{TP}}{\mathrm{TP} + \mathrm{FP} + \mathrm{FN}}
\end{align}
where $TP$, $FP$ and $FN$ refers to true-positive, false-positive and false-negative, respectively. %
}

\subsection{Regularization with Knowledge Distillation}
Distilling knowledge from old model to the current model has shown promises to mitigate catastrophic forgetting of neural network in a CL setting~\cite{li2017learning, douillard2020podnet}. Here, the old model with the knowledge of already observed tasks acts as a teacher model and the purpose is to distill knowledge from teacher model to student model (\ie, current model) such that prediction of these two models matches for previously seen tasks samples. Knowledge distillation is often performed on the probability space to relate the temperature smoothed probability distribution of old to the new model~\cite{li2017learning, cermelli2020modeling}. Feature distillation on the other hand is performed on the feature space extracted from the intermediate layers of neural network to match corresponding local or global statistics of feature maps between old and new model~\cite{michieli2021continual, michieli2019incremental, douillard2020podnet, douillard2021plop}.

Assume that the extracted feature and the prediction using teacher model from step $t-1$ for input $\mat{X}$ are $\vec{f}^{t-1}$ and $\hat{y}^{t-1}$, respectively. Similarly, let $\vec{f}^{t}$ and $\hat{y}^{t}$ be the feature and prediction respectively for the same input using the student model at step $t$. The knowledge distillation is performed between teacher and student model by minimizing the following loss function
\begin{align}
    \mathrm{L}_{\text{kd}}(\mat{X};\Theta^t)
    \coloneqq - \sum_{c \in {\mathbb{C}^{1:t}}}^{} \hat{y}^{t-1}_c~\log\,(\hat{y}^{t}_c)\;.
    \label{eqn:kd_loss_cl}
\end{align}
Feature distillation strategy applies constraint on the similarity between old and new feature representation and is performed by minimizing a notion of distance between corresponding features of $\vec{f}^{t-1}$ and $\vec{f}^{t}$, such as the distance induced by the $\ell_1$ norm as
\begin{align}
    \text{L}_{\ell{1}}(\mat{X};\Theta^t) \coloneqq  \big\|{\vec{f}^{t-1} - \vec{f}^{t}} \big\|_1\;.
    \label{eqn:l1_loss_kd}
\end{align}

\section{Methodology}
\label{sec:methodology}

In this section, we first discuss our main contribution, \emph{the \underline{S}ubspace \underline{D}istillation (SD)} , and its properties. Next, we will elaborate on how the SD will be used for classification and segmentation problems.

\paragraph{\textbf{Subspace Distillation}}

Let $\vec{F}_{i}^{t} \in \mathbb{R}^{d \times p}$ and $\vec{F}_{i}^{t-1} \in \mathbb{R}^{d \times p}$ be the extracted features from layer $i$ of DNNs at step $t$ and $t-1$, respectively. The goal of feature distillation is %
to ensure that the learned knowledge at step $t-1$ remain unchanged at step $t$ while training DNNs on novel dataset $\mathcal{D}^{t}$. To avoid cluttering equations, from now on we drop the layer index, unless it is not clear from the context. In conventional feature distillation approaches, old feature maps or statistics are directly matched with corresponding new one.
That is, in order to obtain $\vec{f}_{i}^{t-1}, \vec{f}_{i}^{t}$ from $\vec{F}_{i}^{t-1}, \vec{F}_{i}^{t}$ for distillation per~\cref{eqn:l1_loss_kd}, different pooling strategies (\ie., channel, height or width pooling) have been discussed in PODnet~\cite{douillard2020podnet}. %

Our hypothesis here is that distilling geometric structure of feature distribution can enrich the distillation process and will help mitigating the catastrophic forgetting in continual learning. In doing so, we propose to model feature maps $\vec{F}^{t}$ and $\vec{F}^{t-1}$ with a set of subspaces, $\mathbb{S} = \{\mathcal{S}_{j}\}_{j=1}^\tau$. Soon, we will discuss how the set of subspaces will be constructed for the classification and segmentation problems, but for now, we focus on the main idea. Each subspace $\mathcal{S} \in \mathbb{S}$ is represented by its basis as 
$\mathbb{R}^{d \times m} \ni \vec{P}; m \ll d$
with $\vec{P}^\top \vec{P}=\mathbf{I}_m$. Note that, $m \leq p$. Our goal here is to preserve the subspace structure of the intermediate feature maps extracted from different layers of old model to the new one. We argue that, robust knowledge distillation through maintaining similarity across subspaces will be advantageous to improve continual classification/segmentation paradigms. This is achieved by enforcing a constraint on subspace similarity between the old and new model.
To do so, we propose to minimize the distance between corresponding subspace constructed from feature maps of the old and new model. \\
A valid distance between $\mathcal{S}_i$ and $\mathcal{S}_j$ is a distance that is invariant to the choice of 
the basis of the subspace. To be more specific, assume $\vec{P}_i \in \mathbb{R}^{d \times m}$ and $\vec{P}_j \in \mathbb{R}^{d \times m}$ are the basis for $\mathcal{S}_i$ and $\mathcal{S}_j$, \ie,  $\vec{P}_i\vec{P}_i^\top = \vec{P}_j\vec{P}_j^\top = \mathbf{I}_m$. Then a distance between 
$\mathcal{S}_i$ and $\mathcal{S}_j$ should meet : 
\[
d(\mathcal{S}_i, \mathcal{S}_j) = g(\vec{P}_i,\vec{P}_j) = g(\vec{P}_i\vec{R}_i,\vec{P}_j\vec{R}_j)\;,
\]
where $g:\mathbb{R}^{d \times m} \times \mathbb{R}^{d \times m} \to \mathbb{R}_+$ is the distance function and $\vec{R}_i, \vec{R}_j \in \mathcal{O}_m$
with $\mathcal{O}_m$ denoting the orthogonal group.

\noindent
To be precise, given $\vec{P}_i^t$ and $\vec{P}_i^{t-1}$, we opt to minimize the projection metric~\cite{harandi2015extrinsic} as
\begin{align}
    \label{eqn:sd_metric}
    \delta_{p}^2(\vec{P}_i^{t},\vec{P}_i^{t-1}) \coloneqq 
    \left\| \vec{P}_i^{t\top}\vec{P}_i^t - \vec{P}_i^{t-1\top}\vec{P}_i^{t-1} \right\|^2_F = 2m - 2\left\| \vec{P}_i^{t\top}\vec{P}_i^{t-1}\right\|^2_F\;.
\end{align}
The projection metric is a proper distance on Grassmannian and endows intriguing properties, among them, the length of curves on Grassmannian obtained by  $\delta_{p}(\cdot,\cdot)$ 
is simply related to the length obtained by the geodesic distance via a fixed constant~\cite{harandi2015extrinsic}. 
In order to illustrate the rationale behind computing subspace distance using Eq.\eqref{eqn:sd_metric}, we present a trivial example involving the XY plane residing within three-dimensional space ($\mathbb{R}^3$) as a two-dimensional subspace. Consider the XY plane in $\mathbb{R}^3$ as a 2D subspace in 3D space. Both 
 
\begin{minipage}{0.4\textwidth}
\begin{align*}
    \vec{P}_1 & = 
\begin{pmatrix}
1 &0\\
0 &1\\
0 &0\\
\end{pmatrix}
\end{align*}
\end{minipage}%
\begin{minipage}{0.1\textwidth}
and
\end{minipage}
\begin{minipage}{0.5\textwidth}
\begin{align*}
\vec{P}_2 &= 
\begin{pmatrix}
\cos(\pi/4) &-\sin(\pi/4)\\
\sin(\pi/4) &\cos(\pi/4)\\
0 &0\\
\end{pmatrix}
\end{align*}
\end{minipage}
represent the XY plane; hence their distance should be zero. The distance used in Eq.\eqref{eqn:sd_metric}, \ie, $\left\| \vec{P}_i^{t\top}\vec{P}_i^t - \vec{P}_i^{t-1\top}\vec{P}_i^{t-1} \right\|^2_F$ not only satisfies the required conditions, but also closely related to the geodesics on the Grassmannian~\cite{harandi2015extrinsic}.\\
In Eq. ~\eqref{eqn:sd_metric}, the mapping $f:\mathbb{R}^{d \times m} \to \mathbb{R}^{m \times m}; f(\vec{P}) =\vec{P}\vec{P}^\top$ is a diffeomorphism between Grassmannian and the space of  symmetric matrices (positive semidefinite to be adequate). The induced distance $\delta_{p}^2(\vec{P}_i^{t},\vec{P}_i^{t-1}) = \left\| \vec{P}_i^{t\top}\vec{P}_i^t - \vec{P}_i^{t-1\top}\vec{P}_i^{t-1} \right\|^2_F$ is invariant to the action of the orthogonal group, satisfying the requirement of having a Grassmannian distance. 
Furthermore, since $\vec{P}^\top\vec{P}=\mathbf{I}_m$, the distance can be simplified to $\delta_{p}^2(\vec{P}_i^{t},\vec{P}_i^{t-1}) = 2m - 2\left\| \vec{P}_i^{t\top}\vec{P}_i^{t-1}\right\|^2_F$, which is computationally very attractive in comparison to geodesics on Grassmannian that require computing SVD.

Before formulating the SD loss for continual learning, note that in practice and in order to construct the subspace representing feature map, we rely on  matrix decomposition techniques, and in particular on Singular Value Decomposition (SVD). In other words, given $\vec{F}_{i}^{t}$ and $\vec{F}_{i}^{t-1}$, we first apply SVD to attain  
$\vec{P}_i$ and $\vec{P}_i^{t-1}$ and then use them accordingly for distillation. This operation differs from many common operations in deep learning, in the sense that a somewhat complicated matrix operation is involved, hence one may wonder how backpropagation will work in this case.

\noindent {\textbf{Backpropagation through SVD.}}
First note that for $\delta_{p}^2(\vec{P}_i^{t},\vec{P}_i^{t-1})$, we have  
\begin{align}
    \nabla_{\vec{P}} \triangleq \frac{\partial}{\partial \vec{P}_i^{t}} \delta_{p}^2\big(\vec{P}_i^{t},\vec{P}_i^{t-1}\big)
     = \frac{\partial}{\partial \vec{P}_i^{t}} \bigg\{ 2m -2\left\| \vec{P}_i^{t\top}\vec{P}_i^{t-1}\right \|^2_F \bigg\} = 
     -2 \frac{\partial}{\partial \vec{P}_i^{t}} 
    \Tr\Bigg(\Big(\vec{P}_i^{t\top}\vec{P}_i^{t-1}\Big)^{\top}\Big(\vec{P}_i^{t\top}\vec{P}_i^{t-1}\Big)\Biggl) 
    = -4 \vec{P}^{t-1}\Big(\vec{P}^{t-1}\Big)^{\top}\vec{P}^{t}\;.
    \label{eqn:deriv_P_proj_metric}
\end{align}

Please note that the subspace from previous model acts as a teacher for distillation and we only need  the gradient with respect to $\vec{P}_i^{t}$ to update our current model, hence the above derivation. The next step to update the weights of the current model is to obtain the gradient of the loss with respect to $\vec{F}_{i}^{t}$, which requires us to backpropagate $\nabla_{\vec{P}}$ through the SVD operation. 
By applying chain rule (see~\cite{ionescu2015matrix} for backpropagation through matrix operations) ,
\begin{align}
    \Tr \Big( \nabla_{\vec{P}}^\top \vec{P}_i^{t} \Big) =
    \Tr \Big( \nabla_{\vec{F}}^\top \vec{F}_i^{t}\Big)\;,
    \label{eqn:chain_rule_mbp}
\end{align}
with 
\begin{align}
    \nabla_{\vec{F}} \triangleq \frac{\partial}{\partial \vec{F}_i^{t}} \delta_{p}^2\big(\vec{P}_i^{t},\vec{P}_i^{t-1}\big)\;.
\end{align}
Feature map $\vec{F}_i^{t}$ is decomposed as $\vec{F}_i^{t}=\vec{P}_i^{t}\vec{\Sigma}\vec{Q}^{\top}$, where $\vec{F}_i^t \in \mathbb{R}^{d \times p}$, $\vec{P}_i^t \in \mathbb{R}^{d \times m}$, $\vec{\Sigma} \in \mathbb{R}^{m \times p}$, $\vec{Q} \in \mathbb{R}^{p \times p}$ and $m \le p$ with the constraints $\big(\vec{P}_i^t\big)^\top\vec{P}_i^t = \vec{Q}^{\top}\vec{Q} = \mathbf{I} $. It can be shown that (see \ref{app:full_derivation} for the derivation)

\begin{align}
    \label{eqn:svd_diff_final_form}
    \nabla_{\vec{F}} = \vec{D}\vec{Q^{\top}} - \vec{P}_i^t \Big( \big(\vec{P}_i^t\big)^{\top}\vec{D}\Big)_{\text{diag}}\vec{Q}^{\top} - 2\vec{P}_i^t\vec{\Sigma}\Big(\vec{K}^{\top} \circ \big(\vec{D}^{\top}\vec{P}_i^t\vec{\Sigma}\big)\Big)_{\text{sym}}\vec{Q}^{\top}\;,
\end{align}
where    $\vec{K} \in \mathbb{R}^{p \times p}$ is defined as follows
 
\begin{equation}
    \label{eqn:kij}
    \vec{K}_{ij} =    \begin{cases}
    \dfrac{1}{\sigma_i^2 - \sigma_j^2}, &i \neq j \\
    0, &i=j
    \end{cases}
\end{equation}

\begin{align}
    \label{eqn:svd_diff_accompany_matrix_D}
    \vec{D} = \nabla_{\vec{F}}\vec{\Sigma}^{-1}_{m}\;.
\end{align}
Here, $\vec{\Sigma}_{m} \in \mathbb{R}^{m \times m}$ consists of top $m$ rows and first $m$ columns of $\vec{\Sigma}$.
Note that, $\mat{A}_{\text{diag}}$ is the diagonal part of $\mat{A}$ (\ie, all off-diagonal elements are set to $0$).

\subsection{Subspace Distillation for Continual Classification}

For the task of continual classification, we distill the subspace constructed across samples in a mini-batch. For a mini-batch 
$\big(\mathcal{X}_B, \mathcal{Y}_B\big)$ of size $b$, let $\vec{f}_{j} \in \mathbb{R}^d$ be the latent representation for input $\vec{X}_{j} \in \mathcal{X}_B$. The formulation below can be applied to any layer in a DNN and hence we drop the layer index for the sake of simplicity. 
Assume there are $p = \lfloor b/|\mathbb{C}^t|\rfloor$ samples per class in the mini-batch. In SD, we propose to compute \underline{class-wise subspaces} using latent features generated from both old and new model. 
More specifically, from the mini-batch, we form $\{\vec{F}^{t-1}_{k}\}_{k=1}^{|\mathbb{C}^t|}$ and $\{\vec{F}^t_{k}\}_{k=1}^{|\mathbb{C}^t|}$ with $\vec{F}^{t-1}_{k},\vec{F}^t_{k} \in \mathbb{R}^{d\times p}$ by stacking corresponding samples in class $k$ into a matrix (\ie, $\vec{F}^t_{k} = \big[\vec{f}_{k,1},\cdots,  \vec{f}_{k,p}\big]$, where $\mathcal{Y}_B \ni y_{k,i} = k$). We note that stacking ordering is not important in forming $\vec{F}^{t-1}_{k}, \vec{F}^t_{k}$ as we are merely interested in subspace spanned by the samples.
Next, we represent each class $k$ from the teacher and the student model by its low-dimensional subspace $\vec{P}^{t-1}_{k}, \vec{P}^t_{k} \in \mathbb{R}^{d \times m}, m \leq p$. This is achieved by applying thin SVD to $\vec{F}^{t-1}_{k}, \vec{F}^t_{k}$ and picking up the top left singular vectors. With the above, the SD loss is defined as

\begin{align}
    \label{eqn:sd_loss_cl}
    \ell_{\text{SD}}^{\text{CL}}\big(\mathcal{X}_B, \mathcal{Y}_B \big) \coloneqq
    \dfrac{1}{|\mathbb{C}^t|}\sum_{k=1}^{|\mathbb{C}^t|} \Big( 2m -2\left\Vert \vec{P}_k^{t\top}\vec{P}_k^{t-1}\right\Vert^2_F\Big)\;.
\end{align}

\subsection{Subspace Distillation for Continual Semantic Segmentation}

For class-incremental semantic segmentation problem, we propose to compute \underline{subspace across the channel dimension} of intermediate feature map for each sample in a batch. To reduce the memory overhead while performing subspace distillation, instead of computing subspace from all feature maps at a layer, we split the full feature maps into several smaller group $G$ with $p$ channels. To compute a subspace from each  group, we need to form a matrix representation from the  feature map. To do so, for a particular input, we form $\{\vec{F}^{t-1}_{g}\}_{g=1}^{G}$ and $\{\vec{F}^t_{g}\}_{g=1}^{G}$ with $\vec{F}^{t-1}_{g},\vec{F}^t_{g} \in \mathbb{R}^{d\times p}$ from the feature maps. Here, $d=wh$, where $h$ and $w$ denote the height and width of the feature map. 
We encode the geometry of $\vec{F}^{t-1}_{j},\vec{F}^t_{j}$ by low-dimensional subspaces  
$\vec{P}^{t-1}_{g},\vec{P}^t_{g} \in \mathbb{R}^{d \times m}$ via SVD. The subspace distillation loss is defined as

\begin{align}
    \label{eqn:sd_loss_css}
    \ell_{\text{SD}}^{\text{CSS}}\big(\vec{X}\big) \coloneqq
    \dfrac{1}{G}\sum_{g=1}^{G} \Big( 2m -2\left\Vert \vec{P}_g^{t\top}\vec{P}_g^{t-1}\right\Vert^2_F\Big)\;.
\end{align}

\subsection{Class-Incremental Continual Learning using Subspace Distillation}
In class-incremental learning, we maintain a fixed memory to store a subset of samples using reservoir sampling strategy~\cite{reservoir} from prior tasks. The samples in the memory are subsequently used during training the model on a novel task. We compute distillation loss between subspace constructed from latent feature maps extracted by feeding memory samples to old and new model. Afterwards, by minimizing subspace distillation loss combined with classification loss on novel and memory samples, we adapt DNNs model for class-incremental learning.
The  classification loss used in conjunction with the subspace distillation loss is the cross entropy defined  as: 
\begin{align}
    \label{eqn:ce_loss_cl}
    \ell_{\text{CE}}^{\text{CL}}(\mat{X}, y) =
    - \sum_{c \in {C^{1:t}}}^{} y_c~\log~(\tilde{y}^{t}_{c})\;,
\end{align}
where, $y_c$, and $\tilde{y}^{t}_c$ are the true label and predicted probability for class $c$ respectively for the input $X$ and $\tilde{y}^{t} = f_{\Theta^t}(\mat{X})$.
Putting everything together, the overall  loss used to train our model to tackle catastrophic forgetting in continual class incremental learning is
\begin{align}
    \label{eqn:overall_loss_cl}
    \mathcal{L}_{\mathrm{CL}}(\mathcal{D}^t;\Theta^t) \coloneqq 
    \E\nolimits_{(\mat{X},y) \sim \mathcal{D}^{t},~(\mat{X'},y') \sim \mathcal{M}} \Big[
    \ell_{\mathrm{CE}}^{\text{CL}}\big(\mat{X}, y\big)
    + \alpha \ell_{\mathrm{CE}}^{\text{CL}}\big(\mat{X'}, y'\big)
    + \beta \ell_{\mathrm{SD}}^{\text{CL}}\big(\mat{X'},y'\big)
    \Big]\;,
\end{align}
\noindent
where, $\vec{X}$ represents the images belonging to novel task while $\vec{X^{'}}$ represents the examples from the memory buffer. $\alpha$ and $\beta$ are hyper-parameter used to control the contribution of the second loss term and subspace distillation, respectively. We present the overall steps of training a continual learning model using subspace distillation in \cref{alg:cl}.

\begin{algorithm}[H] 
\caption{Class-Incremental Learning using Subspace Distillation}
\label{alg:cl}
\begin{algorithmic}[1]
\Require{$\text{Dataset}~\mathcal{D}^t,~\text{Memory}~\mathcal{M},~\text{and Model from step}~t-1,~h_{\Theta^{t-1}} = h_{\text{feat}}^{t-1} \circ h_{\text{cls}}^{t-1}$}
\Ensure{The new model at time $t$ with parameters $\Theta^t$}
    \State {$\text{Initialize }\Theta^t~\text{ with }\Theta^{t-1}$}
    \For {iteration $1$  to  max\_iter}
        \State {Sample a mini batch $(\mathcal{X}_B, \mathcal{Y}_B)$ from ${\mathcal{D}^t}$}
        \State {Sample a mini batch $(\mat{X}', {y}')$ from the memory ${\mathcal{M}}$}

        \State {$ \mathcal{\tilde{Y}}_B \leftarrow h_{\Theta^t}(\mathcal{X}_B)$}
        
        \State $\vec{f}^t \leftarrow h_{\text{feat}}^t(\mat{X}')$
        \State {$\tilde{y}' \leftarrow h_{\text{cls}}^{t}(\vec{f}^t)$}

        \State $\vec{f}^{t-1} \leftarrow h_{\text{feat}}^{t-1}(\mat{X}')$

        \State {Compute Cross Entropy loss, $\ell_{\mathrm{CE}}$ between ground truth $\mathcal{Y}_B$ and prediction $\mathcal{\tilde{Y}}_B$ using Eq. \eqref{eqn:ce_loss_cl}}
        \State {Compute Cross Entropy loss, $\ell_{\mathrm{CE}}$ between ground truth $y'$ and prediction $\tilde{y'}$ using Eq. \eqref{eqn:ce_loss_cl}} 
        \State {Compute Subspace Distillation loss, $\ell_{\text{SD}}^{\text{CL}}$ between $\vec{f}^{t-1}$ and $\vec{f}^{t}$ with Eq. \eqref{eqn:sd_loss_cl}}

        \State {Update $\Theta^t$ by minimizing the overall loss defined based on two cross entropy loss, $\ell_{\mathrm{CE}}$ computed for $\mathcal{X}_B$, and $\mat{X'}$ respectively and subspace distillation loss, $\ell_{\mathrm{SD}}$ as in Eq.~\eqref{eqn:overall_loss_cl}}
    \EndFor
\end{algorithmic}
\end{algorithm}

\subsection{Continual Semantic Segmentation using Subspace Distillation}

Since image pixels belonging to prior classes are labeled as background in CSS, old model is employed for distilling knowledge. The idea of knowledge distillation is a crucial step and widely adapted in preserving prior knowledge for CSS. In CSS, we apply subspace distillation loss at intermediate layers of model 
to maintain consistency in geometric structure of latent features. Additionally, we use output distillation to ensure that the current model mimics the output of prior model. In other words, subspace distillation is combined with classification loss and output distillation loss to train CSS model at any step $t$. %
Bellow, we briefly discuss the classification loss and output distillation loss for CSS.

\subsubsection{Classification Loss}
In semantic segmentation, the cross entropy loss is often applied at each pixel of the output generated by the DNN. However, in CSS and to train the model at time $t$, the background class may include prior classes from already seen tasks. Hence, keeping background shift problem of CSS in mind, we define the cross entropy loss as follows:

\begin{align}
    \label{eqn:ce_loss}
    \ell_{\mathrm{CE}}^{CSS}(\mat{X}, \vec{Y};\Theta^t) =
    -\dfrac{1}{|\mat{X}|} \sum_{(x, y) \in (\mat{X}, \mat{Y})}^{} \sum_{c \in {C^{t}}}~{y}^{t}_{x, c} \log~(\tilde{y}^{t}_{x, c})\;,
\end{align}
\noindent 
where ${y}^{t}_{x, c}$ is the ground truth label at pixel $x$ for class $c$ and $\tilde{y}^{t}_{x, c}$ is defined as follows

\begin{equation}
    \label{eqn:predicted_lbl}
    \tilde{y}^{t}_{x, c} =
    \begin{cases}
    \hat{y}^{t}_{x, c}~\text{if label is not background} \\
    \sum_{c^{'} \in {C^{1:t-1}}}^{} \hat{y}^{t}_{x, c^{'}}~\text{if label is background}.\\
    \end{cases}
\end{equation}
Here, $\hat{y}^{t}_{x, c}$ is the predicted probability for class $c$ at pixel $x$ using model at task $t$.

\subsubsection{Knowledge Distillation}
Distilling knowledge from the old model to current one is crucial to mitigate the catastrophic forgetting and accurate semantic segmentation in continual learning setting. Output distillation by mimicking the predicted probability of old model to the new model has been widely used in the literature of continual learning~\cite{li2017learning, buzzega2020dark}. However, because of background shift problem, conventional knowledge distillation approach cannot be applied directly in the continual semantic segmentation. In this work, similar to MiB method, we follow the masked cross entropy by relating the old model's prediction to the new model's prediction after combining new classes probability with background. Therefore, the adapted output distillation loss is defined as follows:

\begin{align}
    \label{eqn:kd_loss}
    \ell_{\mathrm{KD}}^{CSS}(\mat{X}, \vec{Y}) =
    -\dfrac{1}{|\mat{X}|} \sum_{x \in \mat{X}}^{} \sum_{c \in {C^{1:t-1}}}^{} y^{t-1}_{x, c}~\log(\tilde{y}^{t}_{x, c})
\end{align}

where $y^{t-1}_{x, c}$ is the predicted probability using $f_{\theta^{t-1}}$ for class $c$ at pixel $x$ in image $\mat{X}$. We define  $\tilde{y}^{t}_{x, c}$ as follows:

\begin{equation}
    \label{eqn:predicted_lbl}
    \tilde{y}^{t}_{x, c} =
    \begin{cases}
    y^{t}_{x, c} &\text{if label $c \in {C^{1:t-1}}$ is not background} \\
    \sum_{c^{'} \in {C^{t}}}^{}\limits y^{t}_{x, c^{'}} &\text{if label $c \in {C^{1:t-1}}$ is background}.\\
    \end{cases}
\end{equation}
Since the modified distillation loss does not directly match the predicted background class probability of old model to the new model, the distillation loss plays a vital role in tackling the background label shift problem in class-incremental learning for semantic segmentation.

Finally, the combined loss of our end-to-end continual semantic segmentation method can be written as the linear combination of classification loss, $\ell_{\mathrm{CE}}$, feature distillation using newly proposed subspace distillation loss,$\ell_{\mathrm{SD}}$ , and knowledge distillation loss, $\ell_{\mathrm{KD}}$

\begin{align}
    \label{eqn:overall_loss_css}
    \mathcal{L}_{\mathrm{CSS}}\big(\mathcal{D}^t;\Theta^t\big) \coloneqq 
    \E\nolimits_{(\mat{X}, \vec{Y}) \sim \mathcal{D}^t} \Big[
    \ell_{\mathrm{CE}}^{CSS}\big(\mat{X}, \vec{Y}\big)
    + \alpha \ell_{\mathrm{KD}}^{CSS}\big(\mat{X}, \vec{Y}\big)
    + \beta \ell_{\mathrm{SD}}^{CSS}\big(\mat{X}\big)
    \Big]
\end{align}
Here, $\alpha$ and $\beta$ are the predefined hyperparameter used to control the contribution of $\ell_{\mathrm{KD}}^{CSS}$ and $\ell_{\mathrm{SD}}^{CSS}$ respectively. \cref{alg:css} summarizes the steps needed to be taken to train a new model for CSS.

\begin{algorithm}[H] 
\caption{Subspace Distillation for CSS}
\label{alg:css}
\begin{algorithmic}[1]
\Require{Dataset~$\mathcal{D}^t$,~Model~from~previous~step~$t-1$,~ $h_{\Theta^{t-1}} = h_{\text{encoder}}^{t-1} \circ h_{\text{decoder}}^{t-1}$}
\Ensure{New model at step $t$ with parameter $\Theta^t$}
\State {$\text{Initialize }\Theta^t~\text{ with }\Theta^{t-1}$}

    \For {iteration $1$  to  max\_iter}
        \State {Sample a mini batch $(\mathcal{X}_B, \mathcal{Y}_B)$ from ${\mathcal{D}^t}$}
        
        \State $\vec{F}^t, \hat{\mathcal{Y}_B}^t \leftarrow h_{\Theta^{t}}(\mathcal{X}_B)$
        \State $\vec{F}^{t-1}, \hat{\mathcal{Y}_B}^{t-1} \leftarrow h_{\Theta^{t-1}}(\mathcal{X}_B)$
        
        \State {Compute Cross Entropy loss, $\ell_{\mathrm{CE}}$ between ground truth $\mathcal{Y}_B$ and prediction $\mathcal{\tilde{Y}}_B^t$ using Eq. \eqref{eqn:ce_loss}}

        \State {Compute Output Distillation loss, $\ell_{\mathrm{KD}}$ between the prediction from current and old model,  $\mathcal{\tilde{Y}}_B^{t}$ and $\mathcal{\tilde{Y}}_B^{t-1}$ respectively using Eq. \eqref{eqn:kd_loss}}
        
        \For{$l \gets 1$ to $L$}
            \State {Split $\vec{F}^t[l]$ and $\vec{F}^{t-1}[l]$ into sub group}
            \State {Compute Layer-wise Subspace Distillation loss, $\ell_{\mathrm{SD}}$ between $\vec{F}^{t-1}[l]$ and $\vec{F}^{t}[l]$} using Eq. \eqref{eqn:sd_loss_css}
        \EndFor
        
        \State {Update $\Theta^t$ by minimizing the linear combination of $\ell_{\mathrm{CE}}$, $\ell_{\mathrm{KD}}$ and $\ell_{\mathrm{SD}}$ with Eq.~\eqref{eqn:overall_loss_css}}
    \EndFor
\end{algorithmic}
\end{algorithm}
\section{Experiments}
\label{sec:experiments}

We start this section by describing the datasets, architectures, and implementation details used in our experiments for both continual image classification and semantic segmentation. %
For classification problem we focus on class-incremental and task-incremental settings while in case of continual semantic segmentation problem, we only follow the class-incremental setting. %

\paragraph{\textbf{Continual Learning for Classification}}

We evaluate our proposed method on 3 different benchmark datasets: MNIST~\cite{lecun1998gradient},
CIFAR-10~\cite{krizhevsky2009learning}, %
Tiny Imagenet~\cite{tinyimagenet}. 
Following DER settings, To quantitatively demonstrate the effectiveness of our proposed subspace distillation method in different tasks settings, we split the MNIST, and CIFAR-10 datasets into sequences of 5 tasks having 2 classes per task. Tiny Imagenet is split in 10 tasks with equal number of classes in all sequential tasks (20 classes per task).
In our comparative analysis we consider eight state-of-the-art regularization and distillation methods including LwF, oEWC, SI, iCARL, A-GEM, ER, DER and DER++. In our experiment, we follow class-incremental (CI) and task-incremental (TI) protocols described in~\cite{van2019three}. The identity of task is provided along with input sample in TI setting while in case of CI setting, task identity is absent. In our experiments, we partition data into distinct sets of non-overlapping classes, hence applicable in both class-incremental and task-incremental scenarios. 
Furthermore, we maintain a consistent ordering of all classes across all algorithms, guaranteeing that each algorithm receives identical data for every task. %
Compared to the class-incremental scenario, where task identity is missing, task-incremental learning benefits from having access to task identifiers, which in return helps in selecting appropriate classifiers, rendering it a comparatively easier scenario. Conversely, the class-incremental scenario poses a challenge due to the absence of task identity.

\paragraph{\textbf{Implementation Details}}
Task-incremental learning can be implemented using either a multi-head or single-head classifier, depending on the specific implementation. Following the implementation of DER~\cite{buzzega2020dark}, in our approach, we employed a single-head classifier model to learn in class-incremental scenarios. However, in task-incremental settings, we relied on the output masking technique of the single-head classifier to identify task-specific classes, leveraging the availability of task identity during inference. The output masking technique is used to selectively mask specific outputs of the single-head classifier, depending on the task at hand. This approach allows the single-head classifier to effectively prioritize the relevant outputs for the current task while disregarding the irrelevant ones.
Following the setting described in~\cite{buzzega2020dark, riemer2018learning}, a neural network with 2 fully connected layers of 100 neurons %
are used to extract latent feature for MNIST dataset. %
For CIFAR and Tiny Imagenet datasets, a modified Resnet18-like~\cite{rebuffi2017icarl}  structure is used for  feature extraction. Finally, a single head linear classifier is used for classification across the experiments on MNIST, CIFAR and Tiny Imagenet datasets. We augment both stream and memory samples by applying random crop and horizontal flip for both CIFAR10 and Tiny-Imagenet datasets~\cite{buzzega2020dark}.%

The SGD optimizer is used for training DNN model throughout the learning experiences with keeping flexibility in selection of batch size and mini-batch size for different task setting. Please refer to the appendix for the task-specific values of hyperparameters. Following DER training scheme, we train our model for one epoch at each learning step on MNIST dataset while for relatively complex dataset such as CIFAR10 and Tiny-Imagenet we use 50 and 100 epochs respectively for training.

\paragraph{\textbf{Continual Semantic Segmentation}}

We benchmark our proposed method against state-of-the-art methods with different task settings on Pascal-VOC 2012~\cite{everingham2015pascal} %
dataset. We follow the experimental setup used in~\cite{cermelli2020modeling} for VOC dataset, baseline implementation and metric. Precisely, in our comparative study, we consider eight methods, elastic weight consolidation (EWC)~\cite{kirkpatrick2017overcoming}, learning without forgetting (LwF)~\cite{li2017learning}, Riemannian walk (RW)~\cite{chaudhry2018riemannian}, ILT~\cite{michieli2019incremental}, MiB~\cite{cermelli2020modeling}, SDR~\cite{michieli2021continual}, GIFS~\cite{cermelli2020few} and PLOP~\cite{douillard2021plop}, 

The \textbf{Pascal-VOC 2012 dataset} has 10582 training images and 1449 validation images which is used for testing. Each pixel belongs to 20 foreground classes against background. We use three different tasks settings: 19-1, 15-5 and 15-5s in the experiments. In the first two settings we incrementally add 1 and 5 novel classes on the base model trained on 19 and 15 classes respectively. However, in the 15-5 setting we add 1 class in 5 consecutive tasks while base model remains similar to the 15-5 setting.
In our evaluation of continual semantic segmentation methods, we follow class overlapped setting where task consists of images that may contain classes belonging to future task with a label of background.

\paragraph{\textbf{Implementation Details}} In our implementation, we use a Deeplab-V3~\cite{chen2017rethinking} architecture with  ResNet-101~\cite{he2016deep} as a backbone that is pretrained on Imagenet~\cite{deng2009imagenet}. Following~\cite{bulo2018place}, to reduce required memory, we use inplace activated batch normalization for training our model. We use Stochastic Gradient Decent (SGD) with a momentum of 0.9 and learning decay of 1e-4 to train our model. Following ~\cite{douillard2021plop}, we crop the images to $512 \times 512$ followed by applying random horizontal flip while training our model at each step on Pascal VOC dataset. 
While computing subspace distillation, we construct subspaces at intermediate layers using a group of 32 channels and we consider top 5 subspaces in our distillation strategy. We train our model with a learning rate of .001 from the second task while first task is trained with a higher learning rate of .01 and each task model is trained for 30 epochs with a batch size of 48 distributed over 4 GPU. We use $\alpha=10$ and $\beta=0.01$ while combining output distillation and KD with pixel-wise cross-entropy loss to compute overall loss. For computational efficiency, we employ O1 optimization from Nvidia APEX library\footnote{\url{https://github.com/NVIDIA/apex}} to train with half precision. Finally, we use the validation image set to evaluate our model. Below,  we present the experimental results of our proposed subspace distillation method for both continual learning with classification and continual semantic segmentation problem.

\subsection{Continual Learning}

\begin{table*}[ht]
\vspace{-10mm}
\centering
\resizebox{.85\textwidth}{!}{\begin{tabular}{ccccccc}
\hline
\multirow{2}{*}{Method} & \multicolumn{2}{c}{S-MNIST} & \multicolumn{2}{c}{S-CIFAR-10} & \multicolumn{2}{c}{S-Tiny Imagenet} \\ \cline{2-7} 
      & Task-IL           & Class-IL          & Task-IL       & Class-IL      & Task-IL          & Class-IL          \\ \hline
JOINT     & 99.65      &  97.92     & 98.29  & 92.20  & 82.04     & 59.87      \\
SGD     & 87.15      &  19.90     & 61.02 & 19.61  & 17.93     & 7.79      \\ \hline
LwF~\cite{li2017learning}             & 99.25      &  20.07     & 63.28 & 19.59  & 15.79     & 8.46      \\
oEWC~\cite{schwarz2018progress}            & 99.10      &  20.00     & 68.27 & 19.47  & 19.20     & 7.56      \\
SI~\cite{zenke2017continual}    & 99.07      &  19.97     & 68.05  & 19.46  & 35.97     & 6.58      \\ \hline
    & \multicolumn{5}{c}{Online Data Stream Setting with Tiny Memory (Buffer Size: 100)} & \\ \cline{2-7}

ER~\cite{rolnick2019experience}              & 97.72      &  73.80     & 77.85  & 32.87  & 28.07                  & 5.85                   \\
DER~\cite{buzzega2020dark}             & 98.48      &  77.12    & 80.72  & 32.43 &  27.73                 &  4.26                  \\
 \rowcolor{Gray}    \textbf{Ours (SD)}   & 98.35      &  \textbf{79.37}    &  \textbf{81.65 }             & \textbf{35.1}                 &   \textbf{30.11}               & \textbf{6.05}              \\ 
                        \hline
    &  \multicolumn{5}{c}{Small Memory (Buffer Size: 200)} & \\ \cline{2-7}
iCARL~\cite{rebuffi2017icarl}           & 98.28     &  70.51     & 88.99 & 49.02 &  28.19                &   7.53                \\
ER~\cite{rolnick2019experience}              & 97.86      &  80.43     & 91.19  & 44.79  & 38.17                  & 8.49                   \\
DER~\cite{buzzega2020dark}             & 98.80      &  84.55    & 91.40  & {61.93} &  40.22                 &  11.87                  \\
    \rowcolor{Gray}     \textbf{Ours (SD)}   & 97.71      &  {85.28}    &  \textbf{92.88 }             & 61.85                 &   39.52               & 8.54              \\ 
    \rowcolor{Gray}     DER~\cite{buzzega2020dark} + \textbf{SD}          & \textbf{98.86}      &  \textbf{86.54}    &  {92.07}    &  \textbf{66.12}    &  \textbf{42.63}                 &  \textbf{12.26}       \\ 
                        \hline
    &  \multicolumn{5}{c}{Medium Memory (Buffer Size: 500)} & \\ \cline{2-7}

iCARL~\cite{rebuffi2017icarl}           & 98.81     &  74.55     & 88.22 & 47.55 &  31.55                &   9.38                \\
ER~\cite{rolnick2019experience}              & 98.89      &  86.57     & 93.61  & 57.74  & 48.64                  & 9.99                   \\
DER~\cite{buzzega2020dark}             & 98.84      &  90.54    & 93.40 & 70.51 &  51.78    &  17.75                  \\
     \rowcolor{Gray}    \textbf{Ours (SD)}   & \textbf{ 99.00 }     &  { 89.00}    &  \textbf{94.86 }             & {71.85}                 &   48.60               & 10.03              \\ 
     \rowcolor{Gray}    DER~\cite{buzzega2020dark} + \textbf{SD}            & {98.98}      & \textbf{91.47}      & {94.68}    & \textbf{75.96}     & \textbf{52.74}                  &  \textbf{19.43}      \\

                        \hline
    &  \multicolumn{5}{c}{Large Memory (Buffer Size: 5120)} & \\ \cline{2-7}
iCARL~\cite{rebuffi2017icarl}           & 98.32      &  70.60     & 92.23  & 55.07  & 40.83   &  14.08                  \\
ER~\cite{rolnick2019experience}              &  99.33      &  93.40     & 96.98  & 82.47  & 67.29     & 27.40       \\
DER~\cite{buzzega2020dark}            & 99.29      & 94.9         & 95.43  & 83.81  & 69.50     & 36.73       \\
     \rowcolor{Gray}    \textbf{Ours (SD)}   &  \textbf{ 99.69}    &  \textbf{ 95.90} & \textbf{ 97.18}              &  {84.75}   &   69.13                &   29.32 \\
     \rowcolor{Gray}    DER~\cite{buzzega2020dark} + \textbf{SD}             & 99.44      & 95.33         &  96.77 &  \textbf{86.32}  & 69.47     & \textbf{37.27}       \\
\hline
\end{tabular}}
\caption{\label{tab:acc_benchmark_cl}Average top-1 accuracy on splited MNIST, CIFAR-10 and Tiny Imagenet for 5, 5 and 10 tasks respectively for standard L3 benchmarks. Best values are represented in bold. %
Higher is better. Results for LwF, oEWC, SI, iCARL, ER, and DER are  from~\cite{buzzega2020dark}.%
}
\end{table*}

We analyze the efficacy of subspace distillation in continual learning setting for classification problem and present the result in Table \ref{tab:acc_benchmark_cl}. 
In comparison, we consider two knowledge distillation based methods (LwF~\cite{li2017learning}, iCARL~\cite{rebuffi2017icarl}), two regularization methods (SI~\cite{zenke2017continual}, oEWC~\cite{schwarz2018progress}) and three memory replay based methods (ER~\cite{rolnick2019experience}, DER~\cite{buzzega2020dark}, DER++~\cite{buzzega2020dark}). Additionally, we provide upper and lower bounds, that reflect to jointly training on tasks one by one or on all tasks with Stochastic Gradient Descent (SGD) without any specialized strategy designed for CL.

The results suggest that regularization methods performs poorly on class incremental learning setting as those methods are developed focusing on task incremental learning setting. This observation indicates that regularization towards the set of old parameters is not suitable for tackling catastrophic forgetting because of the local information modeling weight importance~\cite{buzzega2020dark}. Overall, replay based methods outperform regularization methods with big margin across the datasets.
Regularization methods performs good on MNIST datasets for task incremental settings while they perform notably worse on comparatively complex datasets, \eg CIFAR10, Tiny-Imagenet. We observe that our method outperforms state-of-the-art replay base methods in task incremental learning setting on all datasets. Noticeably, for class incremental learning setting on split CIFAR-10 dataset our method performs noticeably better by 15\% and 8\% percentage points compared to iCARL and ER methods, respectively. However, DER method performs slightly better than subspace distillation method as DER relies on storing additional information  (\ie, logits) together with corresponding image and label from past tasks in the memory for the distillation that requires additional memory requirements. We observe that by combining dark experience replay method DER, we can further improve our result. For instance, we can achieve about 2\% and 5\% performance gain on Tiny-Imagenet and CIFAR10 datasets respectively with medium size memory buffer.

We evaluate our model in a more constrained setting where each sample at a task is presented to model once. In other terms, we train our model for a single epoch at each task given a tiny buffer size of 100 samples. Overall, because of the complex setting, continual learning model faces underfitting problem and struggles to mitigate catastrophic forgetting. Replay-based distillation methods also suffer from poor performance when there are insufficient exemplars available, especially for previously observed classes where no memory exemplars exist. For instance, in S-Tiny Imagenet, during the final task, with just 100 memory exemplars, only one example is stored for the 100 previously seen classes out of a total of 180 prior classes, leading to a subpar performance of replay-based distillation methods. We observe that our method considerably improves on baseline ER method across the datasets and performs significantly better than DER. For example, our method enjoys around 2\% improvement on both CIFAR-10 and Tiny-Imagenet datasets than state-of-the-art DER method for class incremental setting. However, we think that this scenario deserves further investigation because of its complex nature.

In task-incremental setting, we notice better performance compared to class-incremental setting for all methods across the datasets. The reason behind such observation is the presence of task identifier at the test time in task-incremental learning that makes the problem easier. We see that SI and oEWC perform badly on relatively complex datasets such as CIFAR10, and Tiny-Imagenet though regularization based methods are designed particularly for task-incremental scenario and our method outperforms both methods with unbridgeable gap in performance. Subspace distillation method performs remarkably better than iCARL across the settings with different memory size. For example, our method outperforms iCARL by 15\% percentage points  with large memory buffer on Tiny-Imagenet, and by 24\% percentage points with medium size memory on CIFAR10 dataset. We also note that our method shows competitive performance with other memory replay based methods. We see around 2\% improvement on ER for task incremental setting on Tiny Imagenet dataset while the performance gain is around 15\% on CIFAR10 dataset for both small and medium size memory. Furthermore, by combining our method with DER, we see 4\%, 5\% and 3\% performance improvements on DER with small, medium and large size memory buffer for task-incremental setting on CIFAR10 dataset.

\subsection{Continual Semantic Segmentation}

Table~\ref{tab:iou_benchmark_voc} reports the IOU of different baseline strategies and subspace distillation method for three scenarios: (i) 19-1, (ii) 15-5 and (iii) 15-1 overlap scenarios on Pascal-VOC dataset. The results suggest that our proposed subspace distillation based method outperforms state-of-the-art methods in 19-1 and 15-5 task settings by a significant margin. Though our method does not match the performance of PLOP in 15-5s task setting, we observe about 13\% IOU improvement on MiB at last task. Furthermore our method outperforms ILT by around 30\% in IOU of last task. We note that subspace distillation shows consistency in retaining prior knowledge across the settings. To show the efficacy of subspace distillation method in tacking catastrophic forgetting, we combine SD with ILT and PLOP methods, and see improvements on IOU in different settings. We observe that by combining SD with ILT the overall performance (\ie, mIOU) of ILT method improves by about 3\% and 2\% for 19-1 and 15-5 task settings respectively while performance remains similar for 15-5s setting. Similarly, we notice 2\% performance improvement of PLOP methods after combining SD for 19-1 and 15-5s settings which suggests that SD imposes additional constraint on distillation strategy using multi-scale POD that helps learning complimentary information and tackling catastrophic forgetting better together. Though our method shows little lower plasticity in case of learning new classes, subspace distillation shows promises in retaining prior knowledge and tackling forgetting old classes.

\begin{table*}[ht]
\centering
\resizebox{\textwidth}{!}{\begin{tabular}{cccccccccc}
\hline
\multirow{2}{*}{Method} & \multicolumn{3}{c}{19-1 (2 tasks)}                   & \multicolumn{3}{c}{15-5 (2 tasks)}                      & \multicolumn{3}{c}{15-1 (6 tasks)} \\ \cline{2-10} 
                        & 0-19 & 20 & \multicolumn{1}{c|}{All} & 0-15 & 16-20 & \multicolumn{1}{c|}{All} & 0-15 & 16-20 & All \\ \hline
 JOINT                       & 77.60     & 76.60   & 77.50     & 79.0     & 72.80      & 77.50      &  79.00    & 72.80      & 77.50 \\

 SGD                       & 6.80     & 12.90   & 7.10     & 2.10     & 33.10      & 9.80      &  0.20    & 1.80      & 0.60 \\
EWC~\cite{kirkpatrick2017overcoming}                       & 26.90     & 14.00   & 26.30     & 24.30     & 35.50     & 27.10  & 0.30      & 4.30      & 1.30    \\
LwF~\cite{li2017learning}                       & 51.20    & 8.50   & 49.10      & 58.90     & 36.60      & 53.30       & 1.00   & 3.90      & 1.80   \\
RW~\cite{chaudhry2018riemannian}                       &  23.30    & 14.20   & 22.90       & 16.60     & 34.90      & 21.20      & 0.00     & 5.20      & 1.30    \\
SDR~\cite{michieli2021continual}                       & 71.30      & 23.40    & 69.0        & 76.30    & 50.20      & 70.10     &  47.30    & 14.70      & 39.50   \\
GIFS~\cite{cermelli2020few}                       & 57.88    & 32.82   & 56.69      & 23.61    & 16.43      & 21.90      & 59.36     & 13.89      & 48.53      \\ \hline
ILT~\cite{michieli2019incremental}                       & 67.75     & 10.88   & 65.05       & 67.08    & 39.23      & 60.45       & 8.75     & 7.99      & 8.56    \\
\rowcolor{Gray} ILT~\cite{michieli2019incremental} + SD                      & \textbf{72.18}     & \textbf{28.02}   & \textbf{70.05}     & \textbf{70.28}    & \textbf{42.63}      &  \textbf{63.70}      &  6.52    &  \textbf{9.03}     & 7.12   \\ \hline
PLOP~\cite{douillard2021plop}                       & 75.35     & 37.35   & 73.54   & 75.73     & 51.71      & 70.09     & 65.12     & 21.11      & 54.64   \\
\rowcolor{Gray} PLOP~\cite{douillard2021plop} + SD                       & \textbf{76.50}     & \textbf{46.39}   & \textbf{75.07}     & 75.63     & 50.17      & 69.57      & \textbf{66.83}     & \textbf{21.56}  & \textbf{56.05}  \\ \hline

MiB~\cite{cermelli2020modeling}                       & 71.43     & 23.59   & 69.15    & 76.37     & 49.97     & 70.08      & 34.22     & 13.50      & 29.29   \\
\rowcolor{Gray} Ours~(MiB~\cite{cermelli2020modeling} + SD)                       &  \textbf{76.09}     & 20.16      & \textbf{73.43}   & \textbf{78.10}       & \textbf{51.21}    & \textbf{71.70}   & \textbf{50.62}     & \textbf{14.41}      &  \textbf{42.00}    \\ \hline

\end{tabular}}
\caption{\label{tab:iou_benchmark_voc}Performance of different continual semantic segmentation method in IOU on Pascal VOC dataset for three continual \textbf{overlap} class learning settings: (i) 19-1 (ii) 15-5 and (iii) 15-1 tasks. Best values are represented in bold. Results for EWC, LwF, RW, ILT, PLOP and MiB are extracted from PLOP~\cite{douillard2021plop} paper while the result for SDR and GIFS are collected from the corresponding original paper. Higher is better. SD: Subspace Distillation.}
\vspace{3mm}

\end{table*}

\begin{figure}[ht]
\vspace{-10mm}
    \centering
        \includegraphics[width=.6\columnwidth]{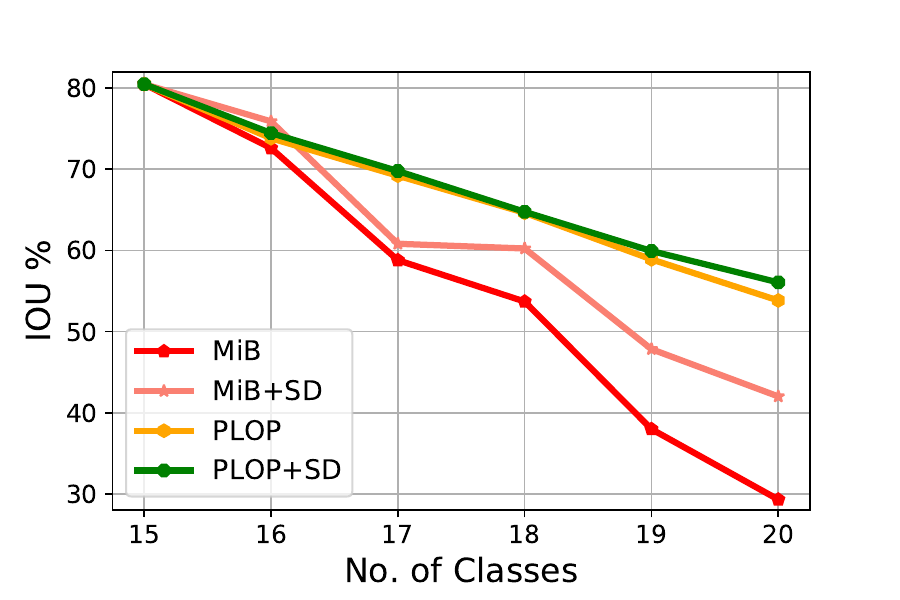}
        \caption{Mean IOU evolution for 15-5s overlap setting on Pascal-VOC dataset.}
        \label{fig:iou_evolution}
\end{figure}

Fig.~\ref{fig:iou_evolution} reports the continual evolution of mean IOU for 15-5s (6 tasks) overlap setting on Pascal-VOC dataset. As depicted in the figure, the performance of MiB drops significantly throughout the learning steps while PLOP shows strength in preserving already learned knowledge and at each learning step PLOP performs significantly better than MiB. We observe considerably improved performance across the learning steps when subspace distillation is added with MiB. SD facilitates MiB to preserve previously learned knowledge and therefore we note about 9\%, and 12\% improvement of mean IOU at task 5, and 6 respectively after adding SD with MiB. Similarly, SD helps PLOP tackling catastrophic forgetting better and we see  improvements on  mean IOU at the end of all learning tasks.

\section{Ablation Studies}
\label{sec:ablation_studies}
In this section, we evaluate our model in terms of changes in activation map, and feature representation in different learning steps in class incremental learning setting on CIFAR10 dataset. Furthermore, in case of continual semantic segmentation on Pascal VOC dataset, we investigate the contribution of each regularizer in subspace distillation strategy and analyze our proposed method with (i) varying dimensionality of subspace and (ii) different no. of channels to construct the subspace.

{\centering
\begin{table*}[ht!]
\addtolength{\tabcolsep}{-5pt}
\resizebox{\textwidth}{!}{
\begin{tabular}{cccccc}
\hline
Input & Method & Task-2 & Task-3 & Task-4 & Task-5 \\
\hline
\multirow{10.65}{*}{\begin{subfigure}{0.2\textwidth}\centering\includegraphics[width=3cm,height=3cm,keepaspectratio]{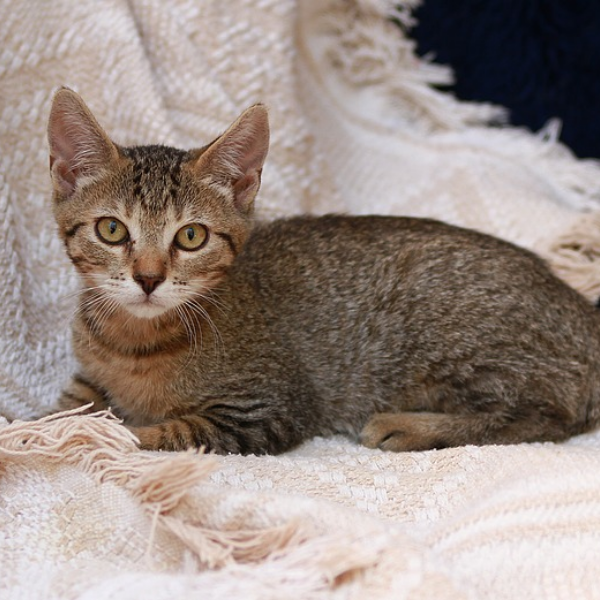}\label{fig:cat}\end{subfigure}} 

& DER &
\begin{subfigure}{0.2\textwidth}\centering\includegraphics[width=3cm,height=3cm,keepaspectratio]{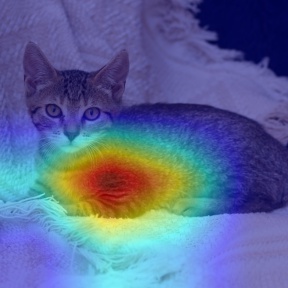}\label{fig:der_cat1}\end{subfigure}&
\begin{subfigure}{0.2\textwidth}\centering\includegraphics[width=3cm,height=3cm,keepaspectratio]{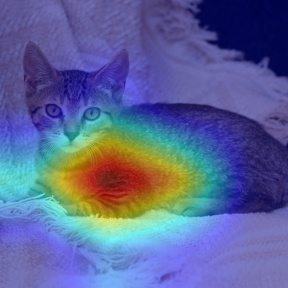}\label{fig:der_cat2}\end{subfigure}&
\begin{subfigure}{0.2\textwidth}\centering\includegraphics[width=3cm,height=3cm,keepaspectratio]{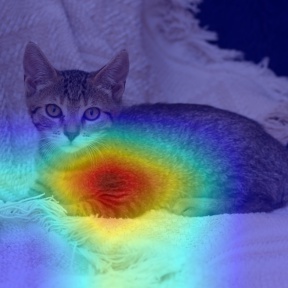}\label{fig:der_cat3}\end{subfigure}&
\begin{subfigure}{0.2\textwidth}\centering\includegraphics[width=3cm,height=3cm,keepaspectratio]{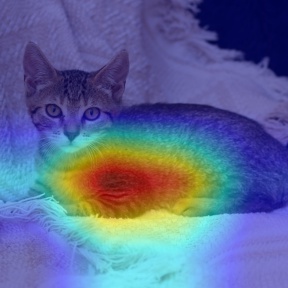}\label{fig:der_cat4}\end{subfigure}\\ \cline{2-6}
\newline
& DER++ &
\begin{subfigure}{0.2\textwidth}\centering\includegraphics[width=3cm,height=3cm,keepaspectratio]{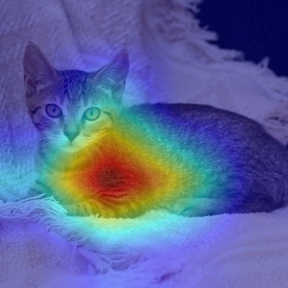}\label{fig:derpp_cat1}\end{subfigure}&
\begin{subfigure}{0.2\textwidth}\centering\includegraphics[width=3cm,height=3cm,keepaspectratio]{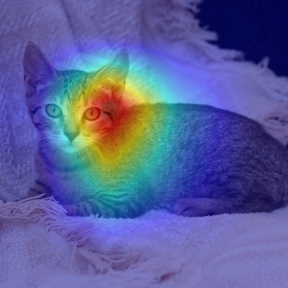}\label{fig:derpp_cat2}\end{subfigure}&
\begin{subfigure}{0.2\textwidth}\centering\includegraphics[width=3cm,height=3cm,keepaspectratio]{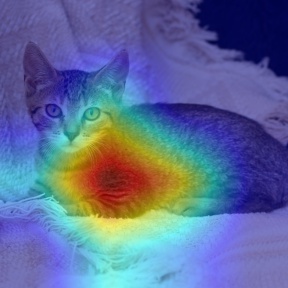}\label{fig:derpp_cat3}\end{subfigure}&
\begin{subfigure}{0.2\textwidth}\centering\includegraphics[width=3cm,height=3cm,keepaspectratio]{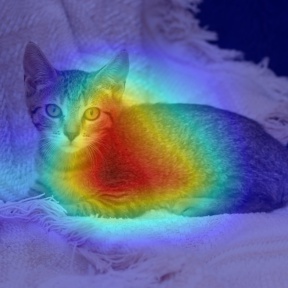}\label{fig:derpp_cat4}\end{subfigure}\\ \cline{2-6}
\newline
& SD &
\begin{subfigure}{0.2\textwidth}\centering\includegraphics[width=3cm,height=3cm,keepaspectratio]{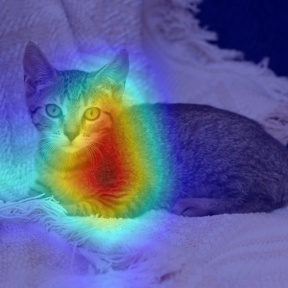}\label{fig:sd_cat1}\end{subfigure}&
\begin{subfigure}{0.2\textwidth}\centering\includegraphics[width=3cm,height=3cm,keepaspectratio]{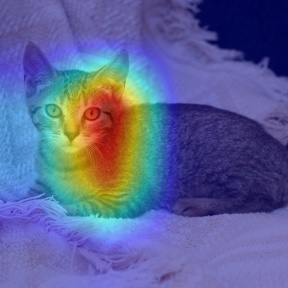}\label{fig:sd_cat2}\end{subfigure}&
\begin{subfigure}{0.2\textwidth}\centering\includegraphics[width=3cm,height=3cm,keepaspectratio]{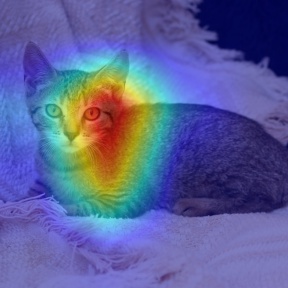}\label{fig:sd_cat3}\end{subfigure}&
\begin{subfigure}{0.2\textwidth}\centering\includegraphics[width=3cm,height=3cm,keepaspectratio]{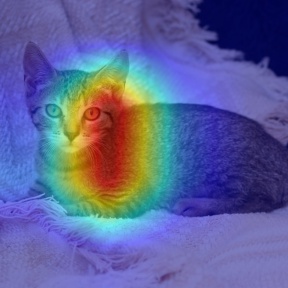}\label{fig:sd_cat4}\end{subfigure}\\ \cline{2-6}

\hline
\multirow{10.65}{*}{\begin{subfigure}{0.2\textwidth}\centering\includegraphics[width=3cm,height=3cm,keepaspectratio]{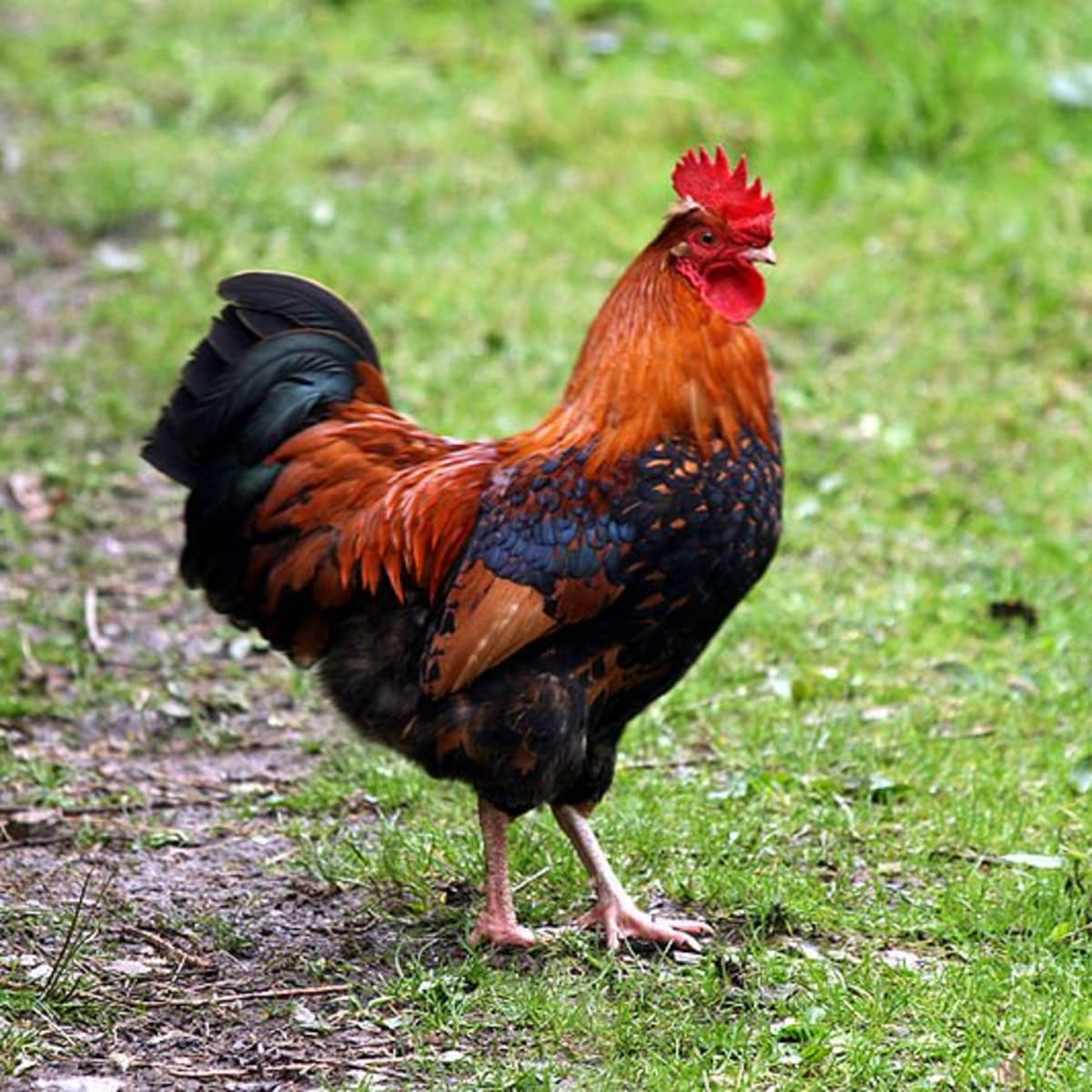}\label{fig:bird}\end{subfigure}} 

& DER &
\begin{subfigure}{0.2\textwidth}\centering\includegraphics[width=3cm,height=3cm,keepaspectratio]{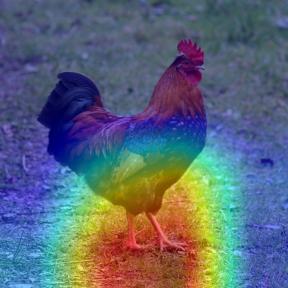}\label{fig:der_bird1}\end{subfigure}&
\begin{subfigure}{0.2\textwidth}\centering\includegraphics[width=3cm,height=3cm,keepaspectratio]{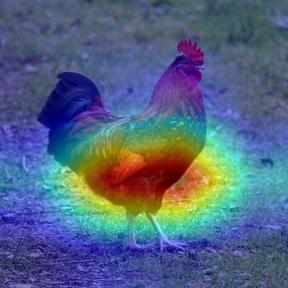}\label{fig:der_bird2}\end{subfigure}&
\begin{subfigure}{0.2\textwidth}\centering\includegraphics[width=3cm,height=3cm,keepaspectratio]{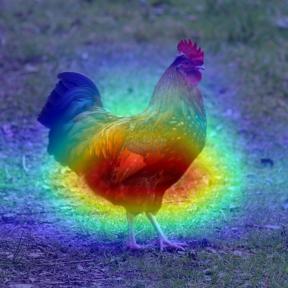}\label{fig:der_bird3}\end{subfigure}&
\begin{subfigure}{0.2\textwidth}\centering\includegraphics[width=3cm,height=3cm,keepaspectratio]{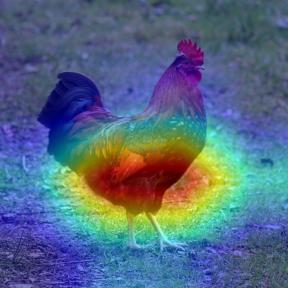}\label{fig:der_bird4}\end{subfigure}\\ \cline{2-6}
\newline
& DER++ &
\begin{subfigure}{0.2\textwidth}\centering\includegraphics[width=3cm,height=3cm,keepaspectratio]{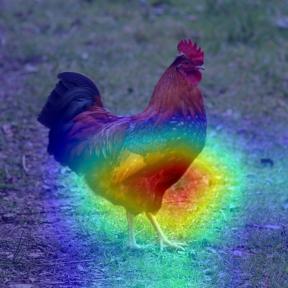}\label{fig:derpp_bird1}\end{subfigure}&
\begin{subfigure}{0.2\textwidth}\centering\includegraphics[width=3cm,height=3cm,keepaspectratio]{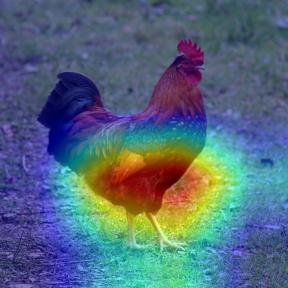}\label{fig:derpp_bird2}\end{subfigure}&
\begin{subfigure}{0.2\textwidth}\centering\includegraphics[width=3cm,height=3cm,keepaspectratio]{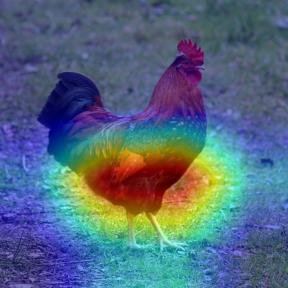}\label{fig:derpp_bird3}\end{subfigure}&
\begin{subfigure}{0.2\textwidth}\centering\includegraphics[width=3cm,height=3cm,keepaspectratio]{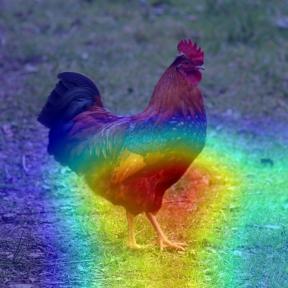}\label{fig:derpp_bird4}\end{subfigure}\\ \cline{2-6}
\newline
& SD &
\begin{subfigure}{0.2\textwidth}\centering\includegraphics[width=3cm,height=3cm,keepaspectratio]{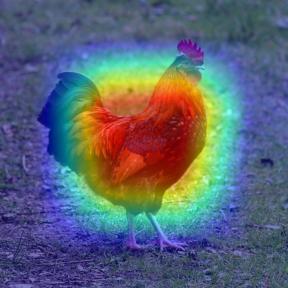}\label{fig:sd_bird1}\end{subfigure}&
\begin{subfigure}{0.2\textwidth}\centering\includegraphics[width=3cm,height=3cm,keepaspectratio]{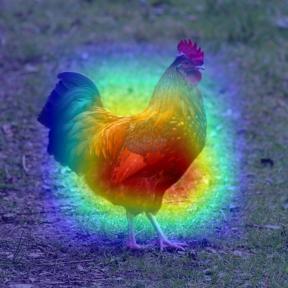}\label{fig:sd_bird2}\end{subfigure}&
\begin{subfigure}{0.2\textwidth}\centering\includegraphics[width=3cm,height=3cm,keepaspectratio]{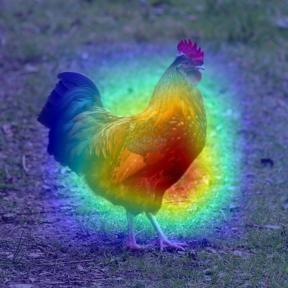}\label{fig:sd_bird3}\end{subfigure}&
\begin{subfigure}{0.2\textwidth}\centering\includegraphics[width=3cm,height=3cm,keepaspectratio]{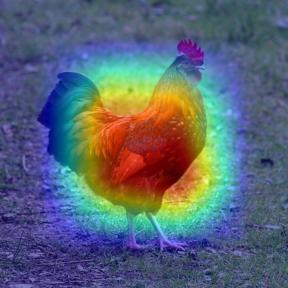}\label{fig:sd_bird4}\end{subfigure}\\

\hline
\end{tabular}
}
\addtolength{\tabcolsep}{1pt} 
\caption{The transition of what neural network is looking at in different learning tasks on CIFAR10 dataset using GradCAM~\cite{jacobgilpytorchcam}. The model trained using our subspace distillation method consistently attends in the regions of the images where the feature of the cat or bird (\ie., around the head or body) is located while the activated regions shift significantly for state-of-the-art DER++ and DER. %
}
\label{tab:grad_cam}
\end{table*}
}

\noindent
\textbf{Changes in Activation Map.}
We examine our model in terms of where neural network is looking at the input images for decision making and whether the activated region has been changed over time in continual learning setting. To do so, We first train a neural network on 5 tasks CIFAR10 datasets and, at the end of learning different tasks, we feed the model with an image that is presented to the model at second step followed by computing heatmap images using GradCAM~\cite{selvaraju2017grad}. We present the result in Figure \ref{tab:grad_cam} showing the evolution of the important regions in the images. The result suggests that DER method activated the body part of cat while our method looks at face region of cat image. Furthermore subspace distillation consistently activates the same regions and the changes in the activation map is lower than the state-of-the-art of DER++ method. Similarly, both DER and DER++ show inconsistency in activating the interested region where bird is located in the image while SD method exhibits robustness in activating the body part of bird in consecutive tasks.

\begin{figure}[htbp!]
    \centering
        \includegraphics[width=.6\columnwidth]{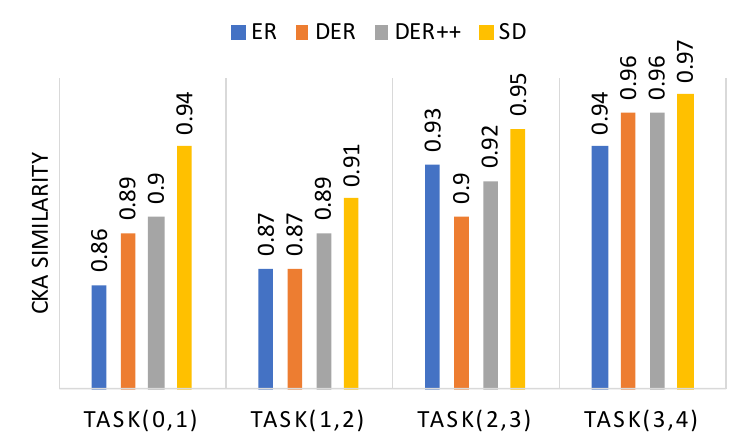}
        \caption{CKA Similarity of the latent feature representation learnt by models in two consecutive task while learning on class incremental CIFAR10 dataset. Subspace Distillation consistently maintains very high representational similarity that is indicative of high capability of knowledge preservation.}
        \label{fig:cka_sim}
\end{figure}

\noindent
\textbf{Similarity in Representation.}
We analyze our models capability of retaining previously learned knowledge by maintaining similar feature representation in the intermediate layer of neural network while learning a series of tasks. Precisely, we employ Centered Kernel Alignment (CKA) metric~\cite{kornblith2019similarity} to compute similarity between intermediate feature representation generated by neural network at task $t$ and $t+1$ on test dataset of CIFAR10 and report the comparative result in Figure \ref{fig:cka_sim}. Overall we observe that the CKA similarity steadily increases as the model faces more tasks and we think that is because the model becomes more stable as it gets adapted incrementally on novel dataset in presence of samples from prior tasks in memory buffer. We notice that our proposed subspace distillation method consistently maintains higher CKA similarity of feature representation throughout the learning experiences. For instance,  SD method outperforms baseline method, ER, on CKA similarity metric by 8\% in the first task while the gap reduces to 3\% in the last task.

\begin{table}[htbh!]
\centering
\resizebox{.65\columnwidth}{!}{\begin{tabular}{ccccccc}
\hline
\multirow{2}{*}{No. of Channels} & \multicolumn{3}{c}{15-5s (6 tasks)}  & \multicolumn{3}{c}{19-1 (2 tasks)} \\ \cline{2-7} 

                        & 0-15 & 16-20 & \multicolumn{1}{c}{All} & 0-19 & 20 & \multicolumn{1}{c}{All}\\ \hline

16             & 66.73     & 21.57   & 55.97  & 76.47     & 46.38   & 75.04\\
32             & 66.83     & 21.56   & 56.05  & 76.50     & 46.39   & 75.07\\
64             & 66.75    & 21.84   & 56.06   & 76.48    & 46.23   & 75.04\\ \hline
\end{tabular}}
\caption{\label{tab:iou_cdim}Mean IOU for 15-5s and 19-1 incremental task setting on Pascal-VOC overlapped dataset with different no. of channels to compute subspace using SVD. %
}
\end{table}

\noindent
\textbf{Varying no. of channels for computing subspace.}
 We evaluate the effect of dimensionality of channels being used to construct the subspace on Pascal VOC dataset for 15-5s and 19-1 tasks setting by combining SD with PLOP. With the increasing no. of channels for constructing subspace, we do not observe any considerable changes in both stability and plasticity as the IOU for new classes and mean IOU for old classes remains stable. As reported in~\ref{tab:iou_cdim}, subspace distillation exhibits robustness against the varying no. of channels used to construct subspace.

\begin{table}[htbh!]
\centering
\resizebox{.5\columnwidth}{!}{\begin{tabular}{cccc}
\hline
\multirow{2}{*}{Dimension of Subspace} & \multicolumn{3}{c}{19-1 (2 tasks)} \\ \cline{2-4} 

                        & 0-19 & 20 & \multicolumn{1}{c}{All}\\ \hline
1             &  76.45    & 46.19   & 75.00\\
3             & 76.49     & 46.43  & 75.06\\
5             & 76.46    & 46.32   & 75.03\\
7             & 76.46    & 46.33   & 75.03\\ \hline
\end{tabular}}
\caption{\label{tab:iou_SD_dim}Mean IOU for 19-1 overlap incremental task setting on Pascal-VOC dataset with different subspace dimension.}
\end{table}

\noindent
\textbf{Subspace Dimensionality.}
To also examine the effect of dimensionality of subspace used in distilling structure for continual semantic segmentation, we combine subspace distillation with PLOP method and report the experimental result on Pascal VOC dataset for 19-1 overlap tasks setting in~\ref{tab:iou_SD_dim}. The results suggest that, with the increase of subspace dimension, the performance on the already observed classes remains similar and subspace dimensionality exhibits less impact on the overall performance.

\noindent
\textbf{Contribution of each Regularizer.} In order to evaluate the contribution of each regularization term in our proposed subspace distillation method for continual semantic segmentation and class-incremental learning, we conducted experiments on the Pascal VOC dataset with varying numbers of tasks and S-CIFAR10 dataset (see~\cref{tab:iou_benchmark_voc_}). Results in \cref{tab:iou_benchmark_voc_1} suggest that both regularizers, $\ell_{\mathrm{KD}}^{CSS}$ and $\ell_{\mathrm{SD}}^{CSS}$, contribute to improving overall performance across the different settings. For example, in the (19-1) 2-task setting, combining regularizer KD with the baseline using CE led to an increase in performance of around 9\%. We also noted an additional 4\% improvement on both previously observed classes and overall performance when regularizer SD was combined with KD in our proposed subspace distillation method.
At the same time, the experimental findings displayed in \cref{tab:iou_benchmark_voc_2} demonstrate that the combination of subspace distillation (SD), $\ell_{\mathrm{SD}}^{CL}$, and memory replay, $\ell_{\mathrm{CE}}^{CL}$ yields substantial improvements in overall accuracy and forgetting on the S-CIFAR10 dataset. Specifically, there is an approximate increase of 9\% in accuracy and a reduction of 15\% in forgetting.

\begin{table}[htbp]
    \centering
    \begin{subtable}{0.44\textwidth}
        \centering
        \resizebox{\textwidth}{!}{\begin{tabular}{ccccccc}
        \hline
        \multicolumn{3}{c}{Method} & \multicolumn{4}{c}{19-1 (2 tasks)} \\ \cline{4-7} 
                CE    &  KD  &  SD  & \multicolumn{2}{c}{0-19 (Old Classes)} & \multicolumn{2}{c}{Overall} \\ \hline
                \cmark   &  \xmark   &   \xmark & \multicolumn{2}{c}{62.30} & \multicolumn{2}{c}{59.85} \\
                \cmark   &  \cmark   &   \xmark &  \multicolumn{2}{c}{71.43 (\textcolor{green}{+9.13})}    &  \multicolumn{2}{c}{69.15 (\textcolor{green}{+9.30})} \\
                    \cmark   &  \cmark   &   \cmark & \multicolumn{2}{c}{\textbf{76.09 (\textcolor{green}{+4.66})}}   & \multicolumn{2}{c}{\textbf{73.43 (\textcolor{green}{+4.28})}}    \\ \hline
        \end{tabular}}
    \caption{\label{tab:iou_benchmark_voc_1} Experimental results of Continual Learning (CL) methods on Pascal VOC dataset for continual (19-1) 2 tasks \textbf{overlap} class learning settings.}
    \end{subtable}
    \hfill
    \begin{subtable}{0.53\textwidth}
        \centering
        \resizebox{\textwidth}{!}{\begin{tabular}{ccccccc}
        \hline
        \multicolumn{3}{c}{Method} & \multicolumn{4}{c}{S-CIFAR10 (5 tasks; C-IL setting) } \\ \cline{4-7} 
        CE    &  Memory Replay  &  SD  & \multicolumn{2}{c}{Avg. Accuracy} & \multicolumn{2}{c}{Forgetting} \\ \hline
        \cmark   &  \xmark   &   \xmark &  \multicolumn{2}{c}{17.32}    & \multicolumn{2}{c}{80.65} \\
        \cmark   &  \cmark   &   \xmark &  \multicolumn{2}{c}{38.97 (\textcolor{green}{+21.65})}    & \multicolumn{2}{c}{43.34 (\textcolor{green}{-37.31})} \\
        \cmark   &  \cmark   &   \cmark &  \multicolumn{2}{c}{\textbf{47.68 (\textcolor{green}{+8.71})}}     &  \multicolumn{2}{c}{\textbf{27.86 (\textcolor{green}{-15.48})}}    \\ \hline
        \end{tabular}}
    \caption{\label{tab:iou_benchmark_voc_2} Experimental results of CL methods trained for one epoch with 500 memory exemplars on the S-CIFAR10 dataset.}
    \end{subtable}
    \vspace{-.25cm}\caption{\label{tab:iou_benchmark_voc_}Experimental results of (a) Continual Semantic Segmentation on Pascal VOC, and (b) Class-Incremenal laerning on S-CIFAR10 indicate that incorporating subspace distillation with memory replay significantly improves the performance of the CL method. }
\end{table}

\noindent
\textbf{Computational Complexity.} Our proposed distillation loss requires computing the subspace basis employing SVD. For an input feature vector $\vec{F} \in \mathbb{R}^{d \times m}; d> m$, computational complexity of SVD is $\mathcal{O}\big(d \times m \times \min(d, m)\big) = \mathcal{O}(dm^2)$. Assuming, $\vec{P}^{t} \in \mathbb{R}^{d \times n}$ and $\vec{P}^{t-1} \in \mathbb{R}^{d \times n}$ be the basis of top $n$ subspaces used in our computation, the computation cost of ${\vec{P}^{t}}^\top\vec{P}^{t-1}$ is $\mathcal{O}(n \times d \times n) = \mathcal{O}(dn^2)$. Therefore, the total computation cost of our proposed subspace distillation loss can be expressed as  $\mathcal{O}(dm^2 + dn^2)$. During our experiments on the 5-tasks CIFAR10 dataset, we found that one iteration of our replay-based subspace distillation (SD) method requires approximately 60ms on Tesla P100-SXM2 GPU for a batch of 32 samples, while the replay-based vanilla SGD method requires about 37ms. We rely on the assumption that a limited buffer memory is available throughout the learning process to store exemplars from prior tasks that might restrict the use of our method in privacy-focused applications.

\section{Conclusion}
\label{sec:conclusion}
In this work, we propose a generalised end-to-end continual learning framework where subspace distillation is at the core of it. Here, we model low dimensional intermediate feature representations %
using subspaces. By imposing constraint on maintaining similar subspace between old and new model, we ensure robustness in the model towards catastrophic forgetting. Proposed subspace distillation is equally effective for classification and semantic segmentation problem in continual learning scenarios. Empirical analysis shows that our proposed framework with subspace distillation achieves state-of-the-art performance in multiple settings on Pascal-VOC dataset for continual semantic segmentation and MNIST, CIFAR10, and Tiny-Imagenet datasets for class-incremental classification problem. In future, we would like to investigate more about the efficient use of subspace distillation for long tasks setting for CSS as well as for complex continual learning setting when model is trained on data stream that is presented to model once at training time.

\section*{Acknowledgments}
P.M. and K.R. gratefully acknowledge co-funding of the project by the CSIRO's Machine Learning and Artificial Intelligence Future Science Platform (MLAI FSP). K.R. also acknowledges funding from the CSIRO's ResearchPlus Postgraduate Scholarship. M.H. gratefully acknowledges the support from the Australian Research Council (ARC), project DP230101176.

\newpage
\bibliographystyle{elsarticle-num}
\bibliography{egbib.bib}
\newpage
\appendix

\section{Proof of Eq.~\ref{eqn:svd_diff_final_form}}
\label{app:full_derivation}

For the sake of simplicity, we avoid feature map index $i$ and task identifier $t$ in the following proof.
Let, feature map $\mat{F}$ be a ${d \times p}$ matrix and $d \ge p$. Then, $\mat{F}$ can be decomposed with $\mat{F}=\mat{P}\mat{\Sigma}\mat{Q}^{\top}$, where $\mat{P} \in \mathbb{R}^{d \times m}$, $\mat{\Sigma} \in \mathbb{R}^{m \times p}$, and $\mat{Q} \in \mathbb{R}^{p \times p}$ such that $ \mat{P}\mat{P}^{\top} = \mat{Q}\mat{Q}^{\top} = \mat{I} $. The differential of $\mat{P}$ can be expressed as
\begin{align}
    \label{eqn:a_dx}
    \dd \mat{F} = \dd \mat{P}\mat{\Sigma}\mat{Q}^{\top} + \mat{P}\dd\mat{\Sigma}\mat{Q}^{\top} + \mat{P}\mat{\Sigma}\dd\mat{Q}^{\top}
\end{align}

The differential $\dd \mat{\Sigma}$ is diagonal like $\mat{\Sigma}$ while $\dd \mat{P}$ and $\dd \mat{Q}$ maintain orthogonality constraints: $\dd \mat{P}^{\top} \mat{P} + \mat{P}^{\top} \dd \mat{P} = 0$ and $\dd \mat{Q}^{\top} \mat{Q} + \mat{Q}^{\top} \dd \mat{Q} = 0$ respectively. By applying the orthogonality of $\mat{P}$ and $\mat{Q}$, Eq.~\ref{eqn:a_dx} can be written as
\begin{align}
    \mat{P}^{\top}\dd \mat{F}\mat{Q} = \mat{P}^{\top}\dd \mat{P}\mat{\Sigma} + \dd\mat{\Sigma} + \mat{\Sigma} \dd\mat{Q}^{\top} \mat{Q}
\end{align}
Since both $\mat{P}^{\top}\dd \mat{P}$ and $\dd\mat{Q}^{\top} \mat{Q}$ are anti-symmetric as well as zero diagonal whereas $\dd \mat{\Sigma}$ is diagonal, $\dd \mat{\Sigma}$ can be written as
\begin{align}
    \label{eqn:a_ds}
    \dd\mat{\Sigma} = \Big(\mat{P}^{\top}\dd\mat{F}\mat{Q}\Big)_{\text{diag}}
\end{align}
with $\mat{A}=\mat{P}^{\top}\dd \mat{P}$, $\mat{B}=\dd\mat{Q}^{\top} \mat{Q}$ and $\mat{R} = \mat{P}^{\top}\dd \mat{F}\mat{Q}$ The off-diagonal part satisfies the following 
\begin{align}
    \label{eqn:a_bij}
    \mat{A} \mat{\Sigma} + \mat{\Sigma} \mat{B} = \mat{R} - \mat{R}_{\text{diag}} \nonumber \\
    \Rightarrow \mat{\Sigma}^{\top}\mat{A}\mat{\Sigma} + \mat{\Sigma}^{\top} \mat{\Sigma} \mat{B} = \mat{\Sigma}^{\top} \Big(\mat{R} - \mat{R}_{\text{diag}}\Big) = \mat{\Sigma}^{\top} \bar{\mat{R}} \nonumber\\
    \Rightarrow 
    \begin{cases}
    \sigma_{i}a_{ij}\sigma_{j} + \sigma_{i}^{2}b_{ij} = \sigma_{i}\bar{\mat{R}}_{ij} \nonumber\\
    -\sigma_{j}a_{ij}\sigma_{i} - \sigma_{j}^{2}b_{ij} = \bar{\mat{R}}_{ji}\sigma_{j}
    \end{cases} \nonumber\\
    \Rightarrow
    b_{ij} = \begin{cases}
    \Big(\sigma_{i}^{2} - \sigma_{j}^{2}\Big)^{-1}~\Big(\sigma_{i}\bar{\mat{R}}_{ij} + \bar{\mat{R}}_{ji}\sigma_{j}\Big) ~,~~~~i \ne j \\
    0~,~~~~~~~~~~~~~~~~~~~~~~~~~~~~~~~~~~~~~~~~~~~~~~i = j
    \end{cases}
\end{align}
here $\bar{\mat{R}} = \mat{R} - \mat{R}_{\text{diag}}$ and $\sigma_{i} = \mat{\Sigma_{ii}}$. Using Eq.~\ref{eqn:a_bij}, we can write $\mat{B}$ as follows
\begin{align}
    \mat{B} = \mat{K} \circ \Big( \mat{\Sigma}^{\top} \bar{\mat{R}} + \bar{\mat{R}}^{\top} \mat{\Sigma} \Big) = \mat{K} \circ \Big( \mat{\Sigma}^{\top} \mat{R} + \mat{R}^{\top} \mat{\Sigma} \Big)
\end{align}
where
\begin{align}
    \label{eqn:a_kij}
    \mat{K}_{ij} =
    \begin{cases}
    \dfrac{1}{\sigma_i^2 - \sigma_j^2}, i \neq j \\
    0, i=j
    \end{cases}
\end{align}
Consequently,
\begin{align}
    \label{eqn:a_dq}
    \dd\mat{Q} = 2\mat{Q}~\Bigg(\mat{K}^{\top} \circ \Big(\mat{\Sigma}^{\top}\mat{P}^{\top}\dd\mat{F}\mat{Q}\Big)_{\text{sym}}~\Bigg)
\end{align}

Using Eq.~\ref{eqn:a_ds} and ~\ref{eqn:a_dq} we can obtain $\dd \mat{P}$ from Eq.~\ref{eqn:a_dx}
\begin{align}
\label{eqn:a_dps}
    \dd \mat{P} \mat{\Sigma} = \dd \mat{F} \mat{Q} - \mat{P} \dd \mat{\Sigma} - \mat{P}\mat{\Sigma}\dd \mat{Q}^{\top}\mat{Q} =: \mat{C}
\end{align}
For any block form $\dd \mat{P} = (\dd \mat{P_1}, \dd \mat{P_2})$, the Eq.~\ref{eqn:a_dps} would be satisfied, where $\dd \mat{P_1} \coloneqq \mat{C}\mat{\Sigma}_d^{-1} \in \mathbb{R}^{m \times d}$ and $\dd \mat{P_2} \coloneqq -\mat{P_1}\dd \mat{P_1}^{\top}\mat{P_2} \in \mathbb{R}^{m \times m-d}$ with $\mat{\Sigma}_d$ being the top $d$ rows of $\mat{\Sigma}$. Therefore,
\begin{align}
\label{eqn:a_dp}
    \dd \mat{P} = \Big( \mat{C}\mat{\Sigma}_d^{-1} ~|~ -\mat{P_1}\mat{\Sigma}_d^{-1}\mat{C}^{\top}\mat{P_2}\Big) ,~\mat{C} = \dd \mat{F} \mat{Q} - \mat{P} \dd \mat{\Sigma} - \mat{P}\mat{\Sigma}\dd \mat{Q}^{\top}\mat{Q}
\end{align}
and we have
\begin{align}
    \nabla_{\mat{F}}:\dd\mat{F} = \nabla_{\mat{P}}:\dd\mat{P} + \nabla_{\mat{\Sigma}}:\dd\mat{\Sigma} + \nabla_{\mat{Q}}:\dd\mat{Q}
\end{align}
As we only consider the basis of subspace $\mat{P}$ in the computation of subspace distillation loss, therefore
\begin{align}
\label{eqn:a_dldx1}
    \nabla_{\mat{F}}:\dd\mat{F} = \nabla_{\mat{P}}:\dd\mat{P}
\end{align}

we simplify $\nabla_{\mat{P}}:\dd\mat{P}$ as follows
\begin{align}
    \label{eqn:svd_diff_d_}
    \nabla_{\mat{P}}:\dd\mat{P} = \Big(\nabla_{\mat{P}}\Big)_{1}:\mat{C}\mat{\Sigma}_d^{-1} + \Big(\nabla_{\mat{P}}\Big)_{2}:-\mat{P_1}\mat{\Sigma}_d^{-1}\mat{C}^{\top}\mat{P_2}
\end{align}
As we consider top $d$ subspace in the computation of subspace distillation loss, therefore $\Big(\nabla_{\mat{P}}\Big)_{2}$ is zero. Consequently
\begin{align}
    \label{eqn:a_svd_diff_d}
    \begin{split}
    \nabla_{\mat{P}}:\dd\mat{P} & = \Big(\nabla_{\mat{P}}\Big)_{1}\mat{\Sigma}_d^{-1}:\mat{C} \\
    & = \underbrace{\Big(\nabla_{\mat{P}}\Big)_{1}\mat{\Sigma}_d^{-1}}_{\mat{D}} : \Big( \dd \mat{F} \mat{Q} - \mat{P} \dd \mat{\Sigma} - \mat{P}\mat{\Sigma}\dd \mat{Q}^{\top}\mat{Q} \Big)  \\
    & = \mat{D}:\dd \mat{F} \mat{Q} ~-~ \mat{D}:\mat{P} \dd \mat{\Sigma} ~-~ \mat{D}:\mat{P}\mat{\Sigma}\dd \mat{Q}^{\top}\mat{Q} ,~ \mat{D}=\Big(\nabla_{\mat{P}}\Big)_{1}\mat{\Sigma}_d^{-1}\\
    & = \mat{D}\mat{Q}^{\top}: \dd \mat{F} ~-~ \mat{P}^{\top}\mat{D}: \dd \mat{\Sigma}~-~ \mat{\Sigma}\mat{P}^{\top}\mat{D}\mat{Q}^{\top}:\dd \mat{Q}^{\top} \\
    & = \mat{D}\mat{Q}^{\top}:\dd \mat{F} ~-~ \mat{P}^{\top}\mat{D}: \dd \mat{\Sigma} ~-~ \mat{Q}\mat{D}^{\top}\mat{P}\mat{\Sigma}:\dd \mat{Q} \\
    & = \mat{D}\mat{Q}^{\top}:\dd \mat{F} ~-~ \mat{P}^{\top}\mat{D}: \Big(\mat{P}^{\top}\dd\mat{P}\mat{Q}\Big)_{\text{diag}} ~-~ \mat{Q}\mat{D}^{\top}\mat{P}\mat{\Sigma}:2\mat{Q}~\Biggl(\mat{K}^{\top} \circ \Big(\mat{\Sigma}^{\top}\mat{P}^{\top}\dd\mat{P}\mat{Q}\Big)_{\text{sym}}~\Biggl) \\
    & = \mat{D}\mat{Q}^{\top}:\dd \mat{F} ~-~ \Big(\mat{P}^{\top}\mat{D}\Big)_{\text{diag}}: \mat{P}^{\top}\dd\mat{P}\mat{Q} ~-~ 2\mat{Q}^{\top}\mat{Q}\mat{D}^{\top}\mat{P}\mat{\Sigma}:~\Biggl(\mat{K}^{\top} \circ \Big(\mat{\Sigma}^{\top}\mat{P}^{\top}\dd\mat{P}\mat{Q}\Big)_{\text{sym}}~\Biggl) \\
    & = \mat{D}\mat{Q}^{\top}:\dd \mat{F} ~-~ \mat{P}\Big(\mat{P}^{\top}\mat{D}\Big)_{\text{diag}}\mat{Q}^{\top}: \dd\mat{P} ~-~ 2\Big(\mat{K}^{\top} \circ \mat{D}^{\top}\mat{P}\mat{\Sigma}\Big)_{\text{sym}}:~ \Big(\mat{\Sigma}^{\top}\mat{P}^{\top}\dd\mat{P}\mat{Q}\Big)\\
    & = \mat{D}\mat{Q}^{\top}:\dd \mat{F} ~-~ \mat{P}\Big(\mat{P}^{\top}\mat{D}\Big)_{\text{diag}}\mat{Q}^{\top}: \dd\mat{P} ~-~ 2\mat{P}\mat{\Sigma}\Big(\mat{K}^{\top} \circ \mat{D}^{\top}\mat{P}\mat{\Sigma}\Big)_{\text{sym}}\mat{Q}^{\top}: \dd\mat{P}
    \end{split}
\end{align}
Finally by using Eq.~\ref{eqn:a_svd_diff_d} in Eq.~\ref{eqn:a_dldx1}, we have
\begin{align}
\label{eqn:a_dldx2}
    \nabla_{\mat{F}} = \mat{D}\mat{Q}^{\top} ~-~ \mat{P}\Big(\mat{P}^{\top}\mat{D}\Big)_{\text{diag}}\mat{Q}^{\top} ~-~ 2\mat{P}\mat{\Sigma}\Big(\mat{K}^{\top} \circ \mat{D}^{\top}\mat{P}\mat{\Sigma}\Big)_{\text{sym}}\mat{Q}^{\top}
\end{align}

\section{Notation and Properties}
The following notations have been used in the derivation
\begin{itemize}
    \item Symmetric part of a square matrix $\mat{A}$, $\mat{A}_{\text{sym}} = \dfrac{1}{2}\Big(\mat{A}^{\top} + \mat{A}\Big)$
    \item $\mat{A}_{\text{diag}}$ be $\mat{A}$ with all off-diagonal elements set to $0$.
    \item Element-wise product $\mat{A} \circ \mat{B}$
    \item Colon product $\mat{A}:\mat{B}=Tr(\mat{A}^{\top}\mat{B})$
    \item The following properties of inner product also have been used
    \begin{align}
        \mat{A}:\mat{B}\mat{C} = \mat{A}\mat{C}^{\top}:\mat{B} = \mat{B}^{\top}\mat{A}:\mat{C}\\
        \mat{A}:\mat{B}\circ\mat{C} = \mat{B}\circ\mat{A}:\mat{C} \\
        \mat{A}:\mat{B}_{\text{sym}} = \mat{A}_{\text{sym}}:\mat{B} \\
        \mat{A}:\mat{B}_{\text{diag}} = \mat{A}_{\text{diag}}:\mat{B}
    \end{align}

\end{itemize}

\vspace{7ex}
\section{Hyperparameter}

\subsection{Class-Incremental Learning}

\begin{table*}[htbp]
\centering
\resizebox{.9\textwidth}{!}{\begin{tabular}{ccccccc}
\hline
\multirow{2}{*}{Method} & \multicolumn{2}{c}{S-MNIST} & \multicolumn{2}{c}{S-CIFAR-10} & \multicolumn{2}{c}{S-Tiny Imagenet} \\ \cline{2-7} 
      & Task-IL           & Class-IL          & Task-IL       & Class-IL      & Task-IL          & Class-IL          \\ \hline
    & \multicolumn{5}{c}{Online Data Stream Setting with Tiny Memory (Buffer Size: 100)} & \\ \cline{2-6}
SD           & lr: .01, $\alpha: 8$~, $\beta: .4$     &  lr: .01, $\alpha: 10$~, $\beta: .5$  & lr:.03 , $\alpha:.5 $~, $\beta:.75 $ & lr:.03 , $\alpha:.5 $~, $\beta:.5 $  &  lr: .03, $\alpha: 1$~, $\beta: .25$                 &   lr: .03, $\alpha: 1$~, $\beta: .25$                \\

                        \hline
    &  \multicolumn{5}{c}{Small Memory  (Buffer Size: 200)} & \\ \cline{2-6}
SD           & lr: .03, $\alpha: 4$~, $\beta: .4$     &  lr: .03, $\alpha: 4$~, $\beta: .4$    & lr:.03 , $\alpha:.4 $~, $\beta:.4 $ & lr:.03 , $\alpha:.4 $~, $\beta:.4 $ &  lr: .03, $\alpha: .1$~, $\beta: .1$                 &   lr: .03, $\alpha: .1$~, $\beta: .1$                 \\
                        \hline
    &  \multicolumn{5}{c}{Medium Memory  (Buffer Size: 500)} & \\ \cline{2-6}

SD           & lr: .1, $\alpha: 1$~, $\beta: .1$    &  lr: .1, $\alpha: 1$~, $\beta: .1$    & lr:.03 , $\alpha:1 $~, $\beta:1 $ & lr:.03 , $\alpha:1 $~, $\beta:1 $ &  lr: .03, $\alpha: .1$~, $\beta: .1$        &   lr: .03, $\alpha: .1$~, $\beta: .1$                \\

                        \hline
    &  \multicolumn{5}{c}{Large Memory  (Buffer Size: 5120)} & \\ \cline{2-6}
SD           & lr: .1, $\alpha: 1$~, $\beta: 2$      &  lr: .1, $\alpha: 1$~, $\beta: 2$     & lr:.03 , $\alpha:1 $~, $\beta:2 $  & lr:.03 , $\alpha:1 $~, $\beta:2 $  &  lr:.03, $\alpha:1$, ~$\beta:2.5 $   &  lr:.03, $\alpha:1$, ~$\beta:2.5 $                  \\
\hline
\end{tabular}}
\caption{\label{tab:hyperparam}Hyperparameter selected for SD method for class-incremental classification.
}
\end{table*}

\vspace{-.2cm}
\subsection{Continual Semantic Segmentation}

\begin{table*}[htbp]
\centering
\begin{tabular}{c|cc|cc|cc}
\hline
Method & \multicolumn{2}{c|}{19-1 (2-Tasks)} & \multicolumn{2}{c|}{15-5 (2-Tasks)} & \multicolumn{2}{c}{15-1 (5-Tasks)} \\ \hline
SD           & \multicolumn{2}{c|}{lr: .001, $\alpha: 10$~, $\beta: .01$ }    &  \multicolumn{2}{c|}{lr: .001, $\alpha: 10$~, $\beta: .01$ } & \multicolumn{2}{c}{lr:.001 , $\alpha:10 $~, $\beta:.01 $ }              \\
                        \hline
\end{tabular}
\caption{\label{tab:hyperparam}Hyperparameter selected for SD method for Continual Semantic Segmentation.
}
\end{table*}

\end{document}